\documentclass{iopjournal}

\usepackage{subcaption}
\usepackage{hyperref}   

\usepackage[numbers,square]{natbib}
\usepackage{booktabs}
\usepackage{multirow}
\usepackage{float}
\usepackage{amssymb}
\usepackage{tabularx}
\usepackage{longtable}
\usepackage{lineno}
\usepackage{tikz}
\usepackage{amsmath}
\usepackage{cleveref}   

\usetikzlibrary{positioning, arrows.meta, calc, shapes.geometric}

\begin{document}

\articletype{Paper} 

\title{Emulating the Forced Response of Climate Models with Generative Machine Learning}

\author{Graham Clyne$^{1,*}$, Julia Kaltenborn$^2$, Peer Nowack$^3$, Claire Monteleoni$^1$, Anasatase Charantonis$^1$}

\affil{$^1$INRIA, Paris, France}

\affil{$^2$MILA, Montreal, Canada}
\affil{$^3$Karlsruhe Institute of Technology (KIT), Karlsruhe, Germany}

\affil{$^*$Author to whom any correspondence should be addressed.}

\email{clyne.graham@gmail.com}


\begin{abstract}
Global climate models are essential tools to simulate past and potential future pathways of climate change, as well as associated climate impacts. 
Shared Socioeconomic Pathways (SSPs) describe a range of future scenarios of global economic and demographic development. These SSPs are intrinsically linked to changes in climate forcings—the external drivers, such as greenhouse gas and aerosol emissions, which change Earth’s energy balance over time. These forcings act as boundary conditions in Earth System Models (ESM), providing insight into the potential climatic impacts of each SSP. 

Running an ESM, however, is extremely computationally expensive, conflicting with the need for large ensemble runs to provide robust estimates in the presence of internal variability and scenario uncertainty. Machine Learning emulators provide a promising avenue towards fast and cheap scenario generation, but until recently lacked uncertainty quantification and the ability to condition on external forcings. Here, we build upon recent work \cite{clyneArchesClimateProbabilisticDecadal2025} and extend it by training on multiple SSPs. We successfully generate scenarios of IPSL-CM6A-LR unseen during training and remain physically consistent with the underlying climate model, even under strong extrapolation scenarios. Our emulator is validated against MESMER-M, a statistical emulator of land surface temperature. Our research demonstrates that our model, ArchesClimate-SSP, does not simply imitate scenarios seen during training, but is actually capable of modeling the response of a climate state to diverse forcings. This is an important step towards reliable and rapid climate model scenario generation.
\end{abstract}

\section{Introduction}
Humanity is facing the consequences of a changing climate, ranging from vanishing ecosystem services (benefits that humans derive from the natural environment) to increased exposure to extreme events, putting the lives of humans, flora, and fauna at risk \cite{portnerTechnicalSummary2022}. To move forward, we need to mitigate, adapt, and develop our future sustainably \cite{fieldClimateResilientPathwaysAdaptation2014}.
A cornerstone of climate change adaptation is humanity's capacity to project the future of our climate under different scenarios. Policymakers need to understand how different emission plans might affect Earth's climate to make informed decisions on behalf of the groups they represent. For decades, scientists have contributed to understanding Earth's future climate by developing climate models \cite{chenFramingContextMethods2021}. Climate models started by modeling only the atmosphere and ocean, but recently they have developed into Earth System Models (ESMs), i.e. they simulate different parts of the Earth in tandem, using physical equations that formalize what we know about how the atmosphere, ocean, cryosphere and biosphere operate and interact \cite{manabeThermalEquilibriumAtmosphere1967, manabeClimateCalculationsCombined1969}. ESMs are run for many different ``what-if'' experiments, helping scientists better understand the Earth's climate \cite{chenFramingContextMethods2021,eyringOverviewCoupledModel2016}. One set of such experiments is called ScenarioMIP \cite{oneillScenarioModelIntercomparison2016}, which compares the climate outputs of different ESMs under different future scenarios. These experiments essentially ask ``how will our Earth's climate look like if humanity behaves in a certain way?''. Those future scenarios of human behaviour are called Shared Socioeconomic Pathways (SSPs) \cite{vanvuurenSharedSocioeconomicPathways2017}. They describe the future in terms of economic, societal, and demographic changes over the coming decades and as such also formulate assumptions about future emissions of greenhouse gases and aerosols. SSPs have further implications for the secondary formation of other forcing agents through atmospheric chemical reactions, in particular tropospheric and stratospheric concentrations of ozone and concerning the atmospheric oxidizing capacity (relevant for the degradation of greenhouse gases such as methane). Changes in anthropogenic greenhouse gases and aerosols expected from each SSP are provided through \textit{Input4MIPs}, a coordinated activity that provides standardized, version-controlled datasets, providing time-dependent emission maps or atmospheric concentration fields that are then used as inputs for the ESMs participating in ScenarioMIP \cite{meinshausenSharedSocioeconomicPathway2020}. In summary, we can, for example, input a future emission scenario into an ESM to simulate its long-term climate response for a given SSP.

The resulting projections have several types of uncertainty. In the near-term, i.e. 30~years, model uncertainty (how unsure we are about the output of the model) is a major component of the the total uncertainty, followed by the climate system's internal variability (the natural, stochastic fluctuations within the climate system), and a small amount of scenario uncertainty (the range of potential future climate outcomes resulting from different plausible pathways of human activity). As the projection moves farther into the future, scenario uncertainty can start to become the dominant uncertainty given a wide range of SSPs \cite{maherMoreAccurateQuantification2021,lehnerPartitioningClimateProjection2020}. 
Unlike model uncertainty, which could be narrowed through improved representations of model physical processes and computing power, scenario uncertainty is irreducible because it pertains to future human agency. Understanding and quantifying this uncertainty is critical for policymakers; it underpins the exploration of strategies that remain effective across various trajectories. By generating different scenarios, researchers can better demonstrate how immediate policy interventions could directly alter the long-term climate. However, ESMs are computationally expensive, restricting the generation of a wide variety of future scenarios.  
This work aims to help climate models explore additional scenarios by providing a robust method to quickly generate spatially and temporally explicit future climate simulations that respond appropriately to a multitude of external forcings. 

In this study, we investigate whether generative Deep Learning (DL) can generate unseen future climate scenarios from a set of external forcings. We also look at the the emulator learns the correct response to each forcing provided. We use ArchesClimate as the architectural framework, an autoregressive probabilistic climate model emulator based on ArchesWeather \cite{clyneArchesClimateProbabilisticDecadal2025,couaironArchesWeatherGenSkillfulComputeefficient2026} and train it on a set of climate projections produced with the IPSL-CM6A-LR ESM \cite{boucherPresentationEvaluationIPSLCM6ALR2020}. We extend the original ArchesClimate framework by conditioning on sets of forcing agents that represent trajectories under different SSP scenarios. We will refer to the extended model generally as ArchesClimate-SSP (AC-SSP). We train our model using monthly averages from 2015-2100 for 8 different future scenarios, in addition to historical runs and pre-industrial control runs. We held out three SSP scenarios for evaluation: SSP4-3.4 and SSP5-8.5 from our validation set, used for architecture and hyperparameter selection, while SSP5-3.4 is a held-out test scenario never used for model selection.

We structure the paper as follows: we first provide a review of the current state of climate model emulation. We then explain the methods used in our research. Following this, we investigate several architectural and training design choices. We show that we can accurately emulate SSP scenarios unseen at training time, compared to a state-of-the-art statistical emulator, MESMER-M \cite{nathMESMERMEarthSystem2022,beuschEmulatingEarthSystem2020}. Next, we investigate the model's ability to learn the impact of each forcing. Finally, we apply AC-SSP to several out-of-distribution tasks, and apply our method to another climate model, CanESM5.

The contributions of this research are as follows: 
\begin{itemize}
    \item Introducing novel techniques to encourage our model to learn the relationship between the external forcings and the climate state.
    \item Providing a DL model that correctly interpolates and extrapolates to the long-term climate statistics of SSPs unseen at training time.
    \item Showing spatial structures of climate variabilty and responses can be better emulated with a training objective that adheres to spatial and temporal structures.
\end{itemize}

\section{Background}

To contextualize our proposed method for accelerating SSP generation, it is necessary to examine previous efforts to improve the computational efficiency of ESMs. We begin by outlining the computational bottlenecks of traditional models, followed by a review of traditional non-DL emulators. We then explore the recent paradigm shift toward machine learning approaches, concluding with the ongoing challenge of incorporating external forcings into these frameworks.

Historically, the primary alternative to ESMs has been the development of reduced-order or intermediate complexity models. These tools trade the complexity of an ESM for computational efficiency, allowing for the rapid simulation of diverse scenarios \cite{leachFaIRv200GeneralizedImpulse2021,meinshausenEmulatingCoupledAtmosphereocean2011,mengisEvaluationUniversityVictoria2020}. To generate spatially explicit data at low computational costs, these emulators are often combined with pattern scaling techniques \cite{wigleyDevelopingClimateScenarios,womackRapidEmulationSpatially2025,tebaldiPatternScalingIts2014} or functions based on historical temperature and precipitation relationships \cite{castruccioStatisticalEmulationClimate2014}. More advanced statistical emulators, such as MESMER and its monthly extension MESMER-M, successfully model internal variability and regional trends for temperature based on global yearly values \cite{beuschEmulatingEarthSystem2020, nathMESMERMEarthSystem2022}. 
The most direct method for improving the speed of ESMs is to increase their computational resources. However, ESMs are already computationally expensive, and the associated engineering and operational costs remain prohibitively high \cite{acostaComputationalEnergyCost2024}. While researchers are increasingly leveraging automatic differentiation and GPU computing to optimize the performance of modern physics-based climate simulations \cite{gelbrechtDifferentiableProgrammingEarth2023,klockeComputingFullEarth2025,lapillonneOperationalNumericalWeather2026}, further research is needed to represent full ESMs, driving the need for alternative modeling approaches.

Deep learning (DL) has emerged as a means to decouple computational performance from physical complexity. While DL applied to weather forecasting has already demonstrated the ability to match or exceed state-of-the-art physics-based models for short-term predictions \cite{aletSkillfulJointProbabilistic2025, priceGenCastDiffusionbasedEnsemble2023,couaironArchesWeatherGenSkillfulComputeefficient2026}, DL applied to climate models aims to translate these gains to decadal and centennial timescales. However, while ML approaches apply well to initial-value modeling problems, constraint-based long-term climate modeling poses novel challenges.

Architectures used to tackle these problems are categorized into deterministic models, which produce a single trajectory from a given initial condition \cite{cresswell-clayDeepLearningEarth2025b,guanLUCIELightweightUncoupled2025}, and generative models, such as diffusion or flow-matching networks, which generate an ensemble of trajectories for a single state \cite{cachayDYffusionDynamicsinformedDiffusion2023,brenowitzClimateBottleGenerative2025}. Generative models create plausible samples that are particularly effective at capturing the stochastic variability inherent in climate systems. While recent models have successfully emulated both isolated ocean components and coupled ocean-atmosphere dynamics, these efforts remain largely focused on stationary climate conditions \cite{wangCoupledOceanAtmosphereDynamics2024,duncanSamudrACEFastAccurate2025,dheeshjithSamudraAIGlobal2025}.

Despite these advancements, capturing the non-stationarity of a long-term forced climate response to external drivers like greenhouse gases and solar irradiation remains a significant hurdle. While some models utilize prescribed sea surface temperatures (SSTs) as inputs, this approach is often criticized as it avoids the fundamental task of simulating how the atmosphere responds to greenhouse gases and aerosols, as the resulting climate trajectory is already largely dictated by the input. There is a notable scarcity of ML emulators capable of prognostic modeling the Earth’s response to Shared Socioeconomic Pathways (SSPs) internally rather than relying on pre-determined ocean conditions, which remains a critical gap for contributing to meaningful policy-making. Early works include \cite{nguyenClimaXFoundationModel2023,watson-parrisClimateBenchV10Benchmark2022,kaltenbornClimateSetLargeScaleClimate2023,lutjensImpactInternalVariability2025}, which use long time series of emissions from forcing agents to predict climate projections for a limited set of climate variables. In contrast to early works, more recent work includes a larger set of variables and autoregressive rollouts; however, they struggle to capture shifting distributions. For example, recent work like ACE2-SOM has explored carbon dioxide (CO$_2$) sensitivity at fixed warming levels (e.g., 2°C or 4°C), but these models often exhibit extreme sensitivity to distribution shifts \cite{clarkACE2SOMCouplingML2025,kochkovNeuralGeneralCirculation2024}. Similarly, AeroGP uses Gaussian processes to capture regional temperature future responses to changing aerosols \cite{deweyAeroGPMachineLearning2025}. While more approaches are beginning to attempt future state generation based on SSPs \cite{immorlanoTechnicalReportUnified2025,mansfieldPredictingGlobalPatterns2020}, accurately modeling the spatiotemporal trajectory of a forced climate remains an open question that our methodology seeks to address.

\section{Methods}
This section introduces the dataset used in this study. We then detail the model architecture and training procedures, concluding with an overview of the metrics and analytical methods used for evaluation.

\subsection{Forcings} \label{sec:forcings}

Our model incorporates forcings from multiple sources, including greenhouse gases, aerosols and ozone. In this work, we incorporate forcings as concentrations and not as emissions. We consider learning the relationship between forcing concentrations and climate states as the first necessary step, and leave direct mappings from emissions to climate states to future work. Here we describe each forcing in detail. Please see \cref{table:forcings} for a summary of the forcings included in AC-SSP.

We incorporate concentrations for three well-mixed greenhouse gases (GHGs) -- carbon dioxide (CO$_2$), methane (CH$_4$), and nitrous oxide (N$_2$O) -- sourced from the \textit{input4MIPs} repository \cite{meinshausenHistoricalGreenhouseGas2017,meinshausenSharedSocioeconomicPathway2020}. These three gases, measured in parts per billion (ppb) for CH$_4$ and nitrous oxide (N$_2$O) and parts per million (ppm) for CO$_2$, were selected as our primary forcing agents due to their dominant roles in the historical anthropogenic greenhouse effect and under typical future SSPs~\cite{canadell2021GlobalCarbon2021}. CO$_2$ is the major anthropogenic greenhouse gas by historical radiative forcing and also under future SSPs, driving long-term climate change due to its longevity and large, persistent increases in emissions. Methane also poses a substantial climate forcing but has a much shorter e-folding time (the time it takes for the concentration of a gas to decrease to approximately 37\% of its original amount) of around 10~years, resulting in potentially faster, scenario-dependent fluctuations in its atmospheric concentrations \cite{canadell2021GlobalCarbon2021}. N$_2$O has an estimated e-folding time of more than 100~years and is the third most important anthropogenic greenhouse gas directly emitted into the atmosphere. These gases are well mixed, meaning they are relatively equally distributed throughout the atmosphere due to their sufficiently long lifetimes for the atmospheric circulation to distribute them across the entire atmosphere.
The GHG concentrations from input4MIPS are at a $0.5$- degree latitudinal monthly resolution, which we regrid to the IPSL model resolution (ca. $2.5\times1.25$ degrees) using first-order conservative regridding. Conservative regridding is a numerical method for transferring data between different grids that ensures the total integral of a quantity, like heat or moisture, is perfectly preserved so that no energy or mass is artificially created or lost.

Next to greenhouse gases, anthropogenic aerosol emissions pose another major human impact on Earth's energy budget, which we consider here. Such aerosols are tiny solid particles or liquid droplets suspended in the atmosphere. We use the atmospheric load for six aerosol species: Aerosol Soluble Nitrate (ASNO3M), Coarse Soluble Nitrate (CSNO3M), Coarse Insoluble Nitrate (CINO3M), Sulfate (SO4), and two black-carbon species, Elemental Carbon Mode Accumulation Insoluble (AIBCM) and Elemental Carbon Mode Accumulation Soluble (ASBCM). In particular sulfate aerosols are known to reflect sunlight, causing a counteracting cooling balance on the historical climate~\cite{niemeierWhatLimitClimate2015}, including important non-linear temperature effects~\cite{feichterNonlinearAspectsClimate2004,mansfieldPredictingGlobalPatterns2020,AerosolClimateSystem2022}. We use data from INCA, a chemistry and aerosol model developed at the Laboratoire des Sciences du Climat et de l'Environnement \cite{lurtonImplementationCMIP6Forcing2020,hauglustaineInteractiveChemistryLaboratoire2004}. INCA is ran beforehand to generate aerosols states that are then used downstream in the IPSL model. We again conservatively regrid this data to a 144x144 resolution from 143x144. Here we simply include the atmospheric load (kg m$^{-2}$) of the aerosols as a rough proxy of Aerosol Optical Depth. 

Finally, we differentiate ozone from the other greenhouse gases, because ozone is formed through chemical reactions in the atmosphere rather than being directly emitted from the Earth's surface \cite{pyleOzoneClimateReview}. In addition, its concentrations differ by orders of magnitude between the troposphere and the stratosphere. This is due to very different ozone production and loss pathways that depend on the availability of sunlight, ozone-depleting substances, and (tropospheric) ozone precursors. For example, in the troposphere, primary ozone production pathways depend on the availability of volatile organic compounds (VOCs) -- including methane -- and of nitrogen oxides (NOx) from both biogenic and anthropogenic sources at Earth's surface. In the stratosphere, ozone production is primarily driven by the Chapman reactions in the presence of high-frequency ultraviolet sunlight, in balance with a variety of major catalytic ozone loss reactions. The latter also depend on anthropogenic emissions of N$_2$O and, famously, on a variety of halogen-containing ozone-depleting substances, which have led to the development of substantial Antarctic ozone holes over past decades \cite{laboratorycslScientificAssessmentOzonea}. In our framework, we use ozone mole fractions from \textit{input4MIPs} covering the troposphere, stratosphere, and even the mesosphere, and conservatively regrid the data from $96\times144$ (latxlon) to $144\times144$ \cite{morgensternOzoneSensitivityVarying2018}. There are $66$ vertical layers ranging from $0.0001$~hPa to $1000$~hPa, but we only include the following layers: $[1000, 850, 500, 200, 100, 50, 10, 1, 0.01, 0.0001]$. We expect ozone to help inform both stratospheric and tropospheric cooling and warming trends \cite{nowackImpactStratosphericOzone2018,maMlozHighlyEfficient2026,sonImpactStratosphericOzone2010,wangExploringOzoneClimate2025a}.

We still include only a subset of the forcings provided to IPSL-CM6A-LR due to project-based computational constraints of this study and our focus on atmospheric dynamics \cite{lurtonImplementationCMIP6Forcing2020}. For example, we assume that the climatic influence of ocean biogeochemical data and land-use cover data was of lower magnitude and could be omitted as a compromise with computational cost and model complexity, and left for future work.

\begin{table}
    \begin{tabularx}{\textwidth}{l X l}

        \multicolumn{3}{c}{\textbf{Forcings}}\\
        \toprule
        Variable Name & Description & Unit \\
        \midrule
        \textit{CO$_2$} & Carbon Dioxide & ppm \\
        \textit{CH$_4$} & Methane & ppb \\
        \textit{N$_2$O} & Nitrous Oxide & ppb \\
        \textit{ASNO3M} & Atmospheric load of nitrate mode accumulation soluble & kg m$^{-2}$ \\
        \textit{CSNO3M} & Atmospheric load of nitrate mode coarse soluble & kg m$^{-2}$ \\
        \textit{CINO3M} & Atmospheric load of nitrate mode coarse insoluble & kg m$^{-2}$ \\
        \textit{SO4} & Atmospheric load of sulfate mode accumulation soluble & kg m$^{-2}$ \\
        \textit{AIBCM} & Atmospheric load of elemental carbon mode accumulation insoluble & kg m$^{-2}$ \\
        \textit{ASBCM} & Atmospheric load of elemental carbon mode accumulation soluble & kg m$^{-2}$ \\
        \textit{Ozone} & Ozone mole fraction, 10 of the 66 available vertical levels (1000, 850, 500, 200, 100, 50, 10, 1, 0.01, 0.0001~hPa) & ppb \\
        \bottomrule
    \end{tabularx}
\caption{External forcings used in AC-SSP. Each forcing is normalized and provided as an input based on its spatial dimensionality. With the exception of atmospheric aerosol loads, which are generated by the INCA chemistry and aerosol model, all forcings are sourced from \textit{input4MIPs}. These data are generated independently of the target IPSL model and are treated as exogenous variables. Forcings are provided as inputs only for the current timestep and are not targets for autoregressive prediction. The physical justification for including each forcing is given in \autoref{sec:forcings}.}\label{table:forcings}
\end{table}

\subsection{Climate Model Data}

We include monthly data from the IPSL-CM6A-LR for eight SSP scenarios, the \textit{historical} scenario, and the \textit{piControl} scenario. Following ScenarioMIP conventions, the scenarios range from an approximate global radiative forcing of 1.9 Wm$^{-2}$ -- a strong climate change mitigation scenario -- to 8.5~Wm$^{-2}$ by 2100. In general, we train our model on all ensemble members (which sample internal variability uncertainty) of each scenario; however, we include only 10 of the 31 available members of the historical run to keep the model focused on future scenarios. 
Please see \autoref{table:scenarios} for the number of members used for each scenario, alongside a high-level description. 

The variables included in AC-SSP are summarized in \autoref{table:variables}. We select this set of variables for three reasons. First, we prioritize variables that represent physical processes governing the climate system rather than surface temperature alone, including surface fluxes ($net\_surface\_flux$, $evspsbl$) and a comprehensive vertical profile of atmospheric dynamics ($hus$, $ta$, $ua$, $va$, $zg$ across $17$ pressure levels), so that the emulator must learn the coupled thermodynamic and dynamic structure of the atmosphere rather than a single diagnostic quantity. Second, the atmospheric variable set is chosen to be comparable with those used by other AI weather and climate emulators \cite{priceGenCastDiffusionbasedEnsemble2023,kochkovNeuralGeneralCirculation2024,couaironArchesWeatherGenSkillfulComputeefficient2026}, enabling more direct comparison of performance and behaviour across the field. Third, we include an oceanic component ($thetao$ across 10 depths) to capture aspects of coupled model dynamics, since IPSL-CM6A-LR is itself a coupled atmosphere-ocean model and surface climate trends are strongly modulated by ocean heat uptake and storage.

For the evaluation of our model, we utilize SSP4-3.4, abrupt-4xCO2, and SSP5-8.5 for validation, and SSP5-3.4 for the test set. We use $historical$, $piControl$, SSP1-1.9, SSP1-2.6, SSP2-4.5, SSP3-7.0 and SSP4-6.0 as training data. SSP5-3.4 offers a valuable testing scenario due to its overshoot profile, in which CO$_2$ concentrations peak mid-century before declining. Overshoot dynamics are becoming increasingly likely in the future, and are therefore important to accurately model \cite{dickauIrreversibleClimateChanges2025}. As it is our only scenario never used for architecture or hyperparameter selection, it is also our sole check against overfitting to the validation set itself. We use SSP5-8.5 and abrupt4xCO2 to evaluate our model's ability to handle out-of-distribution data, a key assessment of our study. Specifically, around the year 2081 -- approximately $66$ years into the dataset -- the CO$_2$ concentrations in SSP5-8.5 exceed the maximum levels seen in the training data (SSP3-7.0 giving the highest CO$_2$ concentration). SSP5-3.4 starts in 2040, and therefore only provides 60~years of data. SSP4-3.4 and SSP5-8.5 start in 2015 and provide 85~years of data. The inclusion of abrupt4xCO2 specifically allows us to assess AC-SSP's ability to distinguish between internal climatic responses and external forced signals. As the experiment is an immediate 4x step of CO$_2$, it requires the model to understand the difference between the CO$_2$ signal and the previous climate state. We hypothesize that this experiment will show if the model is overfitting to the forcing signal.

We select the IPSL-CM6A-LR model \cite{boucherPresentationEvaluationIPSLCM6ALR2020} as our primary data source for several reasons. Its grid resolution ($1.25^\circ \times 2.5^\circ$) provides a computationally efficient yet physically dense dataset ideal for training deep generative emulators. Furthermore, the model’s atmospheric component, LMDZ6, incorporates advanced parameterizations for tropical convection and boundary layer dynamics, providing a state-of-the-art target for our machine learning objectives \cite{hourdinLMDZ6AAtmosphericComponent2020}. A full description of the IPSL scenario ensemble used for training and evaluation, and of the climate variables predicted by AC-SSP, is given in the Appendix (\autoref{sec:datasets}).

\subsection{Architecture}\label{sec:architecture}
For our deep learning framework, we employ AC-SSP, an extension of the ArchesWeather architecture \cite{clyneArchesClimateProbabilisticDecadal2025,couaironArchesWeatherGenSkillfulComputeefficient2026}. ArchesWeather is a Sliding Window (Swin) Vision Transformer based on the Pangu-Weather model \cite{liuSwinTransformerHierarchical2021,biPanguWeather3DHighResolution2022}. It efficiently processes weather/climate states by partitioning spatial regions into tokens, which are then processed through successive attention blocks. These blocks utilize a windowing mechanism that shifts in size and location to capture multi-scale dependencies. As opposed to the methodology of both ArchesWeather and ArchesClimate, which splits the prediction into two tasks with a deterministic and probabilistic component, we instead train the model with an Energy-score loss \cite{AIFSCRPSEnsembleForecasting,langSensitivityMachinelearnedProbabilistic2026}. We found that the single-step optimization of the residual of the detemrinsitic model focused too much on short term variance, and destroyed the longer forced response the deterministic model had learned. The shortcomings of the flow-matching setup are described in \cref{app:flow_destroys_signal}.

We incorporate external forcings through two distinct pathways. Spatially resolved forcings (ozone, aerosols, and spatial GHGs) are, by default, concatenated directly to the input state. In addition to channel-wise concatenation, spatially resolved forcings can be injected via a SPADE-style (SPatially-ADaptive dEnormalization) pathway \cite{parkSemanticImageSynthesis2019}: a lightweight convolutional projector patch-embeds each spatial forcing field to a per-token embedding at the same resolution as the backbone's tokens, which is then added to the token-wise shift and scale parameters of the conditioning (rather than being broadcast as a single global vector) \cite{chenAdaSpeechAdaptiveText2021}.

\subsubsection{Forcing Dropout}
To help learn the forced response, forcing channels are independently dropped at training time. Rather than substituting zeros for a dropped channel, which could otherwise be confused with a forcing that is legitimately near zero, we substitute a learned null token: a per-channel learned constant for spatial forcings. This forces the model to try to learn the role of each masked forcing by asking it to operate without this forcing at training time, thereby asking it to assume what the effect of the forcing would be. 

During training, spatial and non-spatial forcing channels are independently masked and replaced with the learned null token described above. Independent per-channel masking, however, means the model rarely sees every forcing at once. We therefore additionally sample, with probability $p_{\text{all}}$, a training step in which no forcing is masked. We evaluate different values of $p_{\text{all}}$ to determine how often the model needs to see the full set of forcings jointly during training in order to use them correctly at inference time, when they are always fully present; $p_{\text{all}}=0.8$ is the value used by the default model in this research. Please see \cref{app:forcing_dropout_ex} for a concrete example. 

Rather than masking every forcing channel fully independently, physically coupled channels are dropped as a single unit. Two such groups are used: the six aerosol species (\textit{ASNO3M}, \textit{CSNO3M}, \textit{CINO3M}, \textit{SO4}, \textit{AIBCM}, \textit{ASBCM}) and the ten ozone bands described in \autoref{table:forcings}. Every other channel (the three well-mixed greenhouse gases) is masked independently.

Formally, at each training step every channel $c$ (or, for the two physically coupled groups above, every such group jointly) is assigned a binary keep-mask $m_c$ by a two-stage draw:
\begin{equation}
    r \sim \text{Uniform}(0,1), \qquad
    m_c =
    \begin{cases}
        1, & r < p_{\text{all}} \\
        \text{Bernoulli}(1-p_{\text{drop}}), & r \geq p_{\text{all}}
    \end{cases}
\end{equation}
with $r$ drawn once per sample (so all channels are kept together or masked together enters the same branch) and, within the masking branch, each channel's (or group's) Bernoulli draw made independently.

\usetikzlibrary{positioning, arrows.meta, calc, shapes.geometric, fit, backgrounds}

\begin{figure}[htpb]
    \centering
    \resizebox{0.8\linewidth}{!}{%
    \begin{tikzpicture}[
        scale=0.85,
        transform shape,
        >=stealth,
        box/.style={draw, rounded corners=3pt, minimum height=0.8cm, inner sep=2mm, align=center, fill=white, font=\sffamily},
        modebox/.style={box, minimum width=2.4cm, fill=blue!5},
        operator/.style={circle, draw, minimum size=0.6cm, inner sep=0pt, font=\bfseries},
        note/.style={font=\scriptsize\itshape, align=center},
        modegroup/.style={draw, dashed, rounded corners=4pt, inner sep=3mm},
        every path/.style={rounded corners=3pt},
        thick
    ]

    \node[box] (xt) at (0, 7.5) {$X_t$};
    \node[box] (xtm1) at (2.2, 7.5) {$X_{t-1}$};
    \node[box, minimum width=3.2cm] (sf) at (6.0, 7.5) {$F^{\text{spatial}}_t$\\ \footnotesize (GHG, ozone, aerosol)};

    \node[box, minimum width=3.2cm] (dropout) at (6.0, 5.6) {ForcingDropout};

    \node[modebox] (concat) at (3.5, 3.2) {\textbf{Concat} \\ \footnotesize extra channels};
    \node[modebox] (spade) at (6.0, 3.2) {\textbf{SPADE} \\ \footnotesize spatial\_projector};
    \node[modegroup, fit=(concat)(spade)] (modegroup) {};

    \node[operator] (condsum) at (8.5, 1.2) {$\Sigma$};

    \node[box, minimum height=1.4cm, minimum width=4.0cm] (backbone) at (5.5, -1.0) {Backbone};
    \node[operator] (resid) at (5.5, -3.0) {$+$};
    \node[box] (xhat) at (5.5, -4.5) {$\hat{X}_{t+1}$};


    \draw[->] (sf.south) -- (dropout.north);

    \draw[->] (dropout.south) -- (6.0, 4.5) -| (concat.north);
    \draw[->] (dropout.south) -- (spade.north);

    \draw[->] (spade.south) |- (condsum.west);

    \draw[->] (xtm1.south) -- (2.2, 0.0) -| ([xshift=-12mm]backbone.north);
    \draw[->] (concat.south) -- (3.5, 0.0) -| ([xshift=-4mm]backbone.north);
    \draw[->] (condsum.south) -- (8.5, 0.0) -| ([xshift=12mm]backbone.north);

    \draw (xt.south) -- (0, -0.8) coordinate (xt_split);
    \fill (xt_split) circle (2pt); 
    \draw[->] (xt_split) -- ([yshift=2mm]backbone.west);
    \draw[->] (xt_split) |- (resid.west);

    \draw[->] (backbone.south) -- (resid.north);
    \draw[->] (resid.south) -- (xhat.north);

    \end{tikzpicture}%
    }
    \caption{Architecture of the ArchesClimate model configured in a top-to-bottom pipeline. Top: Input preprocessing, \texttt{ForcingDropout}, and spatial conditioning modes. Middle: Aggregation of conditioning tokens ($\Sigma$). Bottom: ArchesClimate backbone receiving state inputs and conditioning.}
    \label{fig:architecture}
\end{figure}

\subsection{Training and Inference}

While training and inference follow the ArchesClimate framework, we implement the following modifications. First, model parameters are updated using an Exponential Moving Average (EMA) to improve stability of long autoregressive generations, and EMA weights are used exclusively for inference. Second, training proceeds in two phases. The first 16000~steps use the Energy/Variogram loss described below, which is sufficient for the model to learn the broad structure of the climate state. Subsequently, we switch to a ``pushforward'' loss to learn the finer structure of the climate state. In this stage, the model is unrolled for multiple steps but gradients are retained only for the final 2~steps. This mechanism forces the model to remain accurate even after several steps of its own compounding error, which we expect to trade a degree of short-horizon precision for long-horizon and extrapolation robustness. We evaluate lengths of $\{2, 4, 8\}$ steps on our validation scenario, SSP4-3.4 (see \autoref{sec:pushforward_search}), and default to $4$ for all subsequent experiments.

\subsubsection{Energy Score Loss}\label{sec:energy_score_loss}

Here we introduce the energy score \cite{gneitingStrictlyProperScoring2007,langSensitivityMachinelearnedProbabilistic2026}, a proper scoring rule that rewards both accuracy and calibrated dispersion. Unlike the traditional per-variable CRPS, which scores each output channel independently and is therefore blind to whether the ensemble reproduces the correct joint structure across variables, levels, and grid points, the energy score is a genuinely multivariate generalization of CRPS and remains sensitive to that cross-dimensional dependence structure. We follow the same fair/unbiased finite-ensemble construction used for CRPS-trained ensemble weather models \cite{aletSkillfulJointProbabilistic2025,kochkovNeuralGeneralCirculation2024,langAIFSCRPSEnsembleForecasting2026}. For a given input, the model produces $M$ stochastic members $\{\hat{y}_1, \dots, \hat{y}_M\}$ by drawing $M$ independent Gaussian noise vectors, each projected through a small learned embedder and added to the token-wise conditioning.

The energy score loss is then the batch-mean of the members' average distance to the target (the \textit{skill} term) minus half their average pairwise distance to each other (the \textit{spread} term):
\begin{equation}\mathcal{L}_{ES} = {\mathbb{E}}\left[\frac{1}{M}\sum_{m=1}^{M} d(\hat{y}_m, y) \; - \; \frac{1}{2M(M-1)}\sum_{i=1}^{M}\sum_{j \neq i} d(\hat{y}_i, \hat{y}_j)\right]\end{equation}
The spread term is subtracted with an $M(M-1)$ (rather than $M^2$) normalization to exclude the zero self-distance terms, giving an unbiased estimator that does not shrink toward zero spread as $M$ grows; at $M=1$ the spread term vanishes and $\mathcal{L}_{ES}$ reduces to the plain latitude-weighted RMSE. We use three members for our energy loss training.

\subsubsection{Variogram Loss}

The energy score's spread term, as defined above, aggregates every variable/level/lat/lon-coordinate of the state into a single scalar distance per batch element. At the dimensionality of a full climate state, this aggregate concentrates tightly around its expectation regardless of how the underlying difference is spatially arranged. Spatially incoherent per-pixel noise and a spatially coherent, physically structured error pattern of the same total variance score almost identically \cite{howardAIForecastEnsembles2026}. To recover sensitivity to spatial structure, we add a Variogram Score term \cite{scheuererVariogramBasedProperScoring2015}, evaluated on pairwise differences at a small, fixed set of spatial lags rather than on the full state at once. For a variable group $v$, spatial axis $\alpha \in \{h,w\}$, and lag $\ell$, define the finite difference at power $p$:
\begin{equation}\delta_{v,\alpha}^{\ell}(x) = \left| x_{v,\alpha+\ell} - x_{v,\alpha} \right|^{p}\label{eq:variogram_diff}\end{equation}
The Variogram score is the mean, over variable groups, axes, and a lag set $\mathcal{L} = \{1,2,4\}$, of the fair (cross-member, bias-corrected) squared discrepancy between the predicted and ground-truth increments:
\begin{equation}\mathcal{L}_{VS} = \underset{v,\alpha,\ell \in \mathcal{L}}{\text{mean}}\left[\frac{1}{M(M-1)}\sum_{i=1}^{M}\sum_{j \neq i} \left(\delta_{v,\alpha}^{\ell}(\hat{y}_i) - \delta_{v,\alpha}^{\ell}(y)\right)\left(\delta_{v,\alpha}^{\ell}(\hat{y}_j) - \delta_{v,\alpha}^{\ell}(y)\right)\right]\end{equation}
As with the energy score's spread term, the cross-member ($i \neq j$) form is an unbiased estimator of $(\mathbb{E}_m[\delta(\hat{y}_m)] - \delta(y))^2$ with no spread-shrinkage incentive; the naive single-member alternative $\text{mean}_m[(\delta(\hat{y}_m) - \delta(y))^2]$ would instead silently reward an artificially tight ensemble. 

In practice, we use an explicit clamp to floor the Variogram score at 0 before it enters the loss. Like the energy score, $\mathcal{L}_{VS}$ is a fair estimator of the mean member-to-truth error minus half the mean member-to-member spread, and a fair/proper scoring rule allows this quantity to go negative, which happens whenever the ensemble's spread exceeds twice its skill. We clamp this at 0 rather than leaving it negative so the term can only ever be stopped from pushing further in an under-dispersed direction. The cost is that this gives up the estimator's strict properness: over-dispersion beyond the skill level is no longer penalized once past this floor, so the Variogram term alone provides no restoring force against excessive spread. In practice we accept this because the energy score's own spread term (\cref{sec:energy_score_loss}) still supplies that restoring force at the global scale; the Variogram term's job is specifically to add spatial-structure sensitivity, not to be the sole regulator of ensemble dispersion.

The total loss is $\mathcal{L} = \mathcal{L}_{ES} + w_{VS} \cdot \mathcal{L}_{VS}$, with $w_{VS}$ ramped linearly from $0$ to its target value ($0.5$) over the first $5000$ steps of Variogram training, so the model first learns to fit the broad state before being penalized on spatial structure.

To perform inference with AC-SSP, we generate climate state time series autoregressively, initialized by IPSL model conditions for a given SSP. For example, to emulate SSP4-3.4, the model is initialized using the first two months of the IPSL scenario (e.g., January and February 2015) and the corresponding forcings. AC-SSP then predicts the subsequent month's state, which is fed back into the model to extend the simulation. We define these generated time series as rollouts, with each rollout serving as a distinct ensemble member. A detailed visualization of this pipeline is available in \cite{clyneArchesClimateProbabilisticDecadal2025,couaironArchesWeatherGenSkillfulComputeefficient2026}.

\begin{table}[h!] 
\label{t1} \centering 
\begin{tabular}{lc}
\toprule
\textbf{Hyperparameter} & \textbf{Value}\\
\midrule
Optimizer & Muon + AdamW (hybrid) \\
Muon Learning Rate & $2e^{-4}$ \\
AdamW Learning Rate & $5e^{-5}$ \\
Schedule & Cosine with Warmup (5000~steps) \\
Warmup Steps & 5000 \\
Betas ($\beta_1, \beta_2$), AdamW group & 0.9, 0.98 \\
Weight Decay & 0.05 \\
\bottomrule
\end{tabular}
\caption{Hyperparameters used in training the deterministic model. Values are reused from ArchesClimate and ArchesWeather. Optimization uses a hybrid scheme: Muon updates every $\geq$2-dimensional weight matrix outside the embedder's input/output boundary layers (the bulk of the backbone), while AdamW updates the embedder boundary layers (patch/de-patch projections, month/hour/timestep embedders, positional axes) and every low-dimensional (bias/norm) parameter, each at its own learning rate; weight decay applies to both groups except AdamW's no-decay (bias/norm) parameters.} \end{table}

\subsubsection{Compute and Running Time} All training and inference reported in this paper run on a single NVIDIA H200~GPU (12~CPU cores, 250GB RAM). For the default $pf=4$ deterministic model, training for $22,000$ steps at batch size $4$ took $6.5$~h wall-clock time. The corresponding generative flow-matching model, also trained for $22,000$ steps at batch size $4$, took $5$~h of wall-clock compute time. The energy-score deterministic model at $pf=4$, trained to $22,000$ steps at batch size $4$, took approximately $6$ hours, a comparable per-step rate to the non-energy-score models above, so the energy-score loss's added cost at $pf=4$ (three stochastic member forward passes during the pushforward window, \cref{sec:architecture_ablations}) is not the dominant factor in this run's time. At inference, generating a full $85$-year ($1020$ monthly steps) autoregressive rollout at our default settings took approximately $80$ seconds for the deterministic model. The energy-score deterministic model's rollout is the same as the deterministic model's, at roughly $1.5$ minutes for a single $85$-year member. 

\subsection{Metrics}\label{sec:metrics}

We use Latitude-Weighted Root Mean Squared Error (RMSE) and Normalized RMSE (for comparison across variables) to evaluate the accuracy of AC-SSP. To evaluate the variance of AC-SSP, we use Interannual Variability (IAV), Ensemble Spread and Spatial Standard Deviation. We use the Hydrostatic Balance as a physical sanity check when comparing ablated models. In meteorology, hydrostatic balance is expressed by combining the Hydrostatic Equation and the Ideal Gas Law \cite{holtonBasicConservationLaws2013}. As a second, independent physical sanity check, we use the Global Energy Balance: we compute the column-integrated atmospheric energy (sensible, potential, kinetic, and latent contributions) and the upper-ocean heat content from the model's own generated state, and check whether their combined rate of change over each rollout month is consistent with the model's own net surface energy flux for that month, as required by conservation of energy \citep{trenberthImperativeClimateChange2009,vonschuckmannHeatStoredEarth2020}; we report the Mean Absolute Error of this residual as a global time series. Please see \cref{app:metrics} for precise definitions of the listed metrics.

\subsection{Ozone and Aerosol Analysis}\label{sec:ozone_aerosol_explanation}

The TCR implies a linear relationship, which cannot be done for ozone and aerosol. For these two groups of forcings, we run emulations of experiments that attempt to isolate both signals. For aerosol, we use $hist--piAer$, which holds aerosol forcings at pre-industrial levels while the rest of the forcings follow $historical$ levels. For ozone, the only available ozone analysis for IPSL (excluding INCA-driven experiments) is $hist--stratO3$ from the Detection and Attribution Model Intercomparison Project (DAMIP). This experiment holds all forcings at pre-industrial levels except for stratospheric ozone, which follows $historical$ levels. We can therefore generate an emulation of the $historical$ experiment (with all forcings included) and compare the difference of the generated $historical$ experiment minus the forcing experiment for both the IPSL and AC-SSP. While it is not ideal that one experiment holds out the forcing and the other experiment holds out everything but the forcing, we are limited by the available data. Both experiments allow us to glimpse into how AC-SSP has learned to incorporate these forcings into long autoregressive generations. We expect ozone held at pre-industrial levels to produce cooling effects and aerosol held at pre-industrial levels to produce warming effects, consistent with their respective roles in the historical forcing budget \cite{laboratorycslScientificAssessmentOzonea}. See \cref{sec:forced_response} for results of these experiments. 

\subsection{Baseline}\label{sec:baseline}
We utilize MESMER-M \cite{nathMESMERMEarthSystem2022,beuschEmulatingEarthSystem2020} as our primary baseline, a statistical climate model emulator designed to produce spatially explicit monthly temperature fields. While the original MESMER framework \cite{beuschEmissionScenariosSpatially2022} focuses on annual temperature maps by accounting for both the long-term local warming signal and grid-point-specific internal variability, MESMER-M extends this capability to a monthly temporal resolution \cite{nathMESMERMEarthSystem2022}. We selected MESMER-M as a benchmark because it is a well-established, non-machine-learning method that decomposes the climate signal into distinct mean and residual components.

The mean component of MESMER-M is modeled as a harmonic Fourier series that represents a smooth seasonal cycle. To account for non-stationarity, both the baseline and the seasonal amplitude are fitted as linear combinations of the annual mean temperature. The remaining residual component, which represents internal variability, is modeled using a first-order autoregressive process.

For our experimental setup, we calibrated MESMER-M using a single member from four different scenarios: SSP1-1.9, SSP1-2.6, SSP2-4.5 and SSP3-7.0, and the historical experiment. A single member suffices here because MESMER-M's two fitted components -- the harmonic seasonal-cycle/forced-response fit and the AR(1) internal-variability model -- are each calibrated against the member's own multi-decadal monthly time series, which already supplies hundreds of monthly samples per scenario; the AR(1) fit is designed to characterize the statistical properties of internal variability in general, not to memorize that one realization's specific noise sequence, so it does not require multiple realizations to be sampled directly at calibration time. Anomalies were calculated relative to the 1850–1900 pre-industrial reference period. During the evaluation phase, we provided MESMER-M with the target annual mean surface temperatures from the IPSL model to predict the monthly temperature of two withheld scenarios, SSP4-3.4 and the overshoot scenario SSP5-3.4. Further technical details regarding the MESMER-M architecture can be found in \cite{nathMESMERMEarthSystem2022}.



\section{Results}

In this section, we evaluate the climate emulations produced by AC-SSP and show that it can accurately capture the spatial and temporal dynamics of different SSP scenarios. All reported results use the training hyperparameters that yielded the highest performance during ablation runs (see \cref{sec:architecture_ablations}). We compare to $MESMER-M$ when possible. Following this statistical validation, we quantify the impact of each forcing on the generated climate states. We then assess the physical consistency of the generated states (\autoref{sec:hydrostatic}). After this, we analyze the SSP5-8.5 scenario and the abrupt4xCO$_2$ scenario, which allows us to test the model's capability for out-of-distribution generalization (\cref{sec:585}). Finally, we follow ClimateSet's \citep{kaltenbornClimateSetLargeScaleClimate2023} reasoning that ML approaches should be tested across different climate models, and test the generalizability of AC-SSP on another climate model (CanESM5).

\subsection{Architectural Ablations} \label{sec:architecture_ablations}

We first fix a set of architectural and training-configuration choices by ablating them independently on the deterministic model, holding all other hyperparameters at their defaults (see \autoref{sec:architecture}). We organize this into two ablation studies, each in its own subsection below: the forcing dropout parameters used during training and the push-forward rollout length.

\begin{table}[ht]
\centering
\resizebox{\textwidth}{!}{%
\begin{tabular}{llcccc}
\toprule
\multirow{2}{*}{\textbf{Ablation}} & \multirow{2}{*}{\textbf{Configuration}} & \textbf{Temporally-Avg.} & \textbf{Spatial Std.} & \textbf{Ensemble} & \textbf{Inter-annua}l \\ 
& & \textbf{RMSE (K)} & \textbf{Deviation} & \textbf{Spread} & \textbf{Variability}
\\ \cmidrule(lr){1-6}
\multirow{4}{*}{Pushforward Length $pf$}
    & $pf=2$ & 1.1831 & 13.5441 & 0.5717 & 0.9394 \\
    & $pf=4$ & 1.1689 & 13.3068 & 0.3865 & 0.6961 \\
    & $pf=8$ & 1.2726 & 13.6955 & 0.5558 & 1.0144 \\
\cmidrule(lr){1-6}
\multirow{6}{*}{Forcing All-Present Probability $p_{\text{all}}$}
    & $p_{\text{all}}=0.0$ & 1.2278 & 13.3151 & 0.3868 & 0.8040 \\
    & $p_{\text{all}}=0.2$ & 1.2226 & 13.3626 & 0.4819 & 0.8566 \\
    & $p_{\text{all}}=0.4$ & 1.2310 & 13.4191 & 0.5561 & 0.8831 \\
    & $p_{\text{all}}=0.6$ & 1.1689 & 13.3068 & 0.3865 & 0.6961 \\
    & $p_{\text{all}}=0.8$ & 0.9941 & 13.3191 & 0.3640 & 0.6124 \\
    & $p_{\text{all}}=1.0$ & 0.9800 & 13.3397 & 0.3476 & 0.6066 \\
\bottomrule
\end{tabular}%
}
\caption{Architectural and training-configuration ablations of the energy-score model, evaluated on the SSP4-3.4 validation scenario. Lower is better for RMSE; IPSL's own spatial standard deviation and inter-annual variability over the same final-decade window are $13.2520$ and $0.6337$ respectively (\autoref{tab:model_performance}), for reference. Each row is a separate training run with all other hyperparameters at their default values (pf=4, $p_{\text{drop}}=0.2$). The RMSE column is temporally averaged: monthly $tas$ is first averaged to annual means, then RMSE (K) is computed against the IPSL target pooled (lat-weighted) over the last 10~years. All models are 5-member ensembles.}
\label{tab:architecture_ablations}
\end{table}

\subsubsection{Multi-step Finetuning} \label{sec:pushforward_search}

We determine the optimal push-forward length ($pf$) by evaluating the energy-score detailed in \autoref{tab:architecture_ablations} (specifically comparing $pf \in \{2, 4, 8\}$, utilizing 5-member ensembles under the SSP4-3.4 scenario). As demonstrated in \autoref{app:seed_stability_energy_score}, this model exhibits sensitivity to initialization; we observe a $22$~\% RMSE spread across three otherwise identical training seeds. Consequently, performance differentials smaller than this baseline variance cannot be reliably distinguished from standard training stochasticity on the basis of single runs. The RMSE values for $pf=2$ and $pf=4$ yield a marginal difference of $0.0142$~K, placing them within this threshold of uncertainty. However, they are distinctly separated by their inter-annual variability (IAV). The IAV for $pf=4$ aligns closely with the IPSL reference target, whereas $pf=2$ significantly overestimates it. This discrepancy is substantial enough to fall well outside the bounds of expected seed noise, strongly favouring the $pf=4$ configuration.

Conversely, $pf=8$ proves unequivocally inferior across all evaluated criteria. It records the highest RMSE and IAV, alongside a spatial standard deviation that diverges furthest from the IPSL baseline. Given that it is both less accurate and more dispersed than its counterparts—deviating by margins that cannot be attributed to noise we discard $pf=8$. We therefore establish \textbf{$pf=4$} as our default configuration for all subsequent analyses.

\subsubsection{Forcing Dropout Parameters} \label{sec:forcing_dropout_search}

This ablation investigates the fundamental trade-off governed by the all-present probability ($p_{\text{all}}$): insufficient exposure to the complete forcing set prevents the model from integrating all forcings jointly, whereas inadequate exposure to partial or missing forcings precludes robust evaluation when forcings are deliberately withheld.

Re-evaluating the six $p_{\text{all}}$ configurations with $pf=4$ using 5-member ensembles (\autoref{tab:architecture_ablations}) reveals a broad improvement in root-mean-square error (RMSE) as $p_{\text{all}}$ increases. Configurations where $p_{\text{all}} \in \{0.0, 0.2, 0.4, 0.6\}$ exhibit an RMSE of approximately $1.17$-$1.23$~K, while the $p_{\text{all}} \in \{0.8, 1.0\}$ configurations achieve superior performance, ranging from $0.98$- $1.03$~K depending on the checkpoint step. Within this top-performing pair, however, the margin is negligible. As established in \autoref{app:seed_stability_energy_score}, this gap falls well within the $22$~\% (approximately $0.22$~K) seed-to-seed variance for this model family, rendering the two configurations indistinguishable on the basis of a single training run's RMSE.

To resolve this ambiguity, we evaluate $p_{\text{all}}=0.8$ and $p_{\text{all}}=1.0$ against physical criteria, employing single-forcing piControl-held attribution ablations referenced against the ERF/TCR estimates (see \cref{app:ap_08_vs_10_forced_response} for a comprehensive per-gas decomposition, and see \cref{app:tcr_explanation} for a definition of the ERF/TCR estimates). Three distinct metrics favour the $p_{\text{all}}=0.8$ configuration. First, it correctly captures the sign of the methane response, whereas the $p_{\text{all}}=1.0$ configuration incorrectly simulates a warming response ($+0.332$~K). Second, although both configurations produce an ozone response of the incorrect sign, the error is substantially attenuated in the $p_{\text{all}}=0.8$ configuration. Finally, while both exhibit significant N$_2$O over-attribution, the magnitude is smaller for $p_{\text{all}}=0.8$ ($\times9.27$ the TCR reference versus $\times12.41$). The sole exception is CO$_2$, where $p_{\text{all}}=0.8$ deviates slightly further from the expected response ($0.129$~K versus $0.063$~K); however, this represents the smallest deviation among the four gases and does not negate the advantages observed elsewhere. Consequently, the physical forcing-attribution evidence robustly supports $p_{\text{all}}=0.8$, despite the aggregate RMSE parity. We thus select \textbf{$p_{\text{all}}=0.8$} as the default checkpoint for AC-SSP and all subsequent analyses in this paper.

To confirm the necessity of the forcing signal, we evaluated a baseline model trained entirely without forcing conditioning. While the autoregressive generation remained numerically stable and captured a generic warming tendency, the state trajectory diverged significantly from the true climate path. We therefore conclude that minimal external forcing information is structurally requisite; the model cannot adequately reconstruct the forced response from the unconditioned state dynamics alone.

\subsection{Comparison to MESMER-M baseline}

Our model demonstrates a high capacity to accurately emulate the IPSL outputs for several SSP experiments. In \autoref{tab:model_performance}, we compare $AC\text{-}SSP$ against the baseline $MESMER\text{-}M$ across several metrics applied to global land surface temperature for the decade $2090$-$2100$. $AC\text{-}SSP$ is a 5-member ensemble, as is $MESMER\text{-}M$; the IPSL reference data provides only a single member for the SSP4-3.4 and SSP5-3.4 scenarios. Performance is evaluated using RMSE for accuracy, spatial standard deviation to quantify geographic temperature gradients (e.g., Arctic vs. Saharan contrasts), Inter-annual Variability (IAV), and ensemble spread to compare internal variability. As there is only a single ground-truth member, we need to be cautious of overstating the performance of RMSE, given the unknown extent of internal variability in the target. All metrics are derived from yearly averages spanning 85~years for SSP4-3.4 and 59~years for SSP5-3.4.

$AC\text{-}SSP$'s RMSE is comparable to $MESMER\text{-}M$'s on both scenarios. On SSP5-3.4, $AC\text{-}SSP$ is in fact the lower-error of the two. Spatial standard deviation also tracks the IPSL target closely for both. $AC\text{-}SSP$'s ensemble spread is consistently larger than $MESMER\text{-}M$'s, consistent with $AC\text{-}SSP$'s autoregressive rollout accumulating more member-to-member divergence over 85~years than $MESMER\text{-}M$'s statistical emulation does. Interestingly, $AC\text{-}SSP$'s IAV is closer to the IPSL target's than $MESMER\text{-}M$'s is, in both scenarios. Because $MESMER\text{-}M$ is conditioned on prescribed global temperatures, its inter-annual variations might be expected to align closely with the IPSL target by construction; that $AC\text{-}SSP$'s IAV tracks the target at least as well suggests the autoregressive rollout is not simply smoothing away real interannual variability, despite having no explicit longer-than-one-step temporal context to draw on. Note that because IPSL provides only a single member, no reference ensemble spread is available for comparison.

\begin{table}[ht]
\centering
\begin{tabular}{llcccc}
\toprule
\textbf{Scenario} & \textbf{Model} & \textbf{RMSE (K)} & \textbf{\shortstack{Spatial Std.\\\\ Deviation}} & \textbf{\shortstack{Ensemble\\\\ Spread}} & \textbf{\shortstack{Inter-annual\\\\ Variability}} \\\\
\midrule
\multirow{3}{*}{\textbf{SSP4-3.4}}
    & $AC\text{-}SSP$ & 0.9843 & 13.3329 & 0.3971 & 0.6161 \\
    & $MESMER\text{-}M$ & 0.9390 & 13.2717 & 0.1179 & 0.7685 \\
    & IPSL & n/a & 13.2520 & n/a & 0.6337 \\
\midrule
\multirow{3}{*}{\textbf{SSP5-3.4}}
    & $AC\text{-}SSP$ & 0.9739 & 13.4434 & 0.3928 & 0.6050 \\
    & $MESMER\text{-}M$ & 1.0963 & 13.6192 & 0.0999 & 0.6375 \\
    & IPSL & n/a & 13.6069 & n/a & 0.5166 \\
\bottomrule
\end{tabular}
\caption{Performance of land surface temperature across metrics and SSP scenarios, for $AC\text{-}SSP$  against $MESMER\text{-}M$ and the IPSL target. All values are based on annual averages of land surface temperature for the decade $2090$-$2100$, as MESMER-M is a land-surface temperature model only. See \autoref{app:metrics} for definitions of the metrics. Ensemble Spread is unavailable for IPSL as it has only one ensemble member for SSP4-3.4 and SSP5-3.4.}
\label{tab:model_performance}
\end{table}

\begin{figure}
\includegraphics[width=1\textwidth]{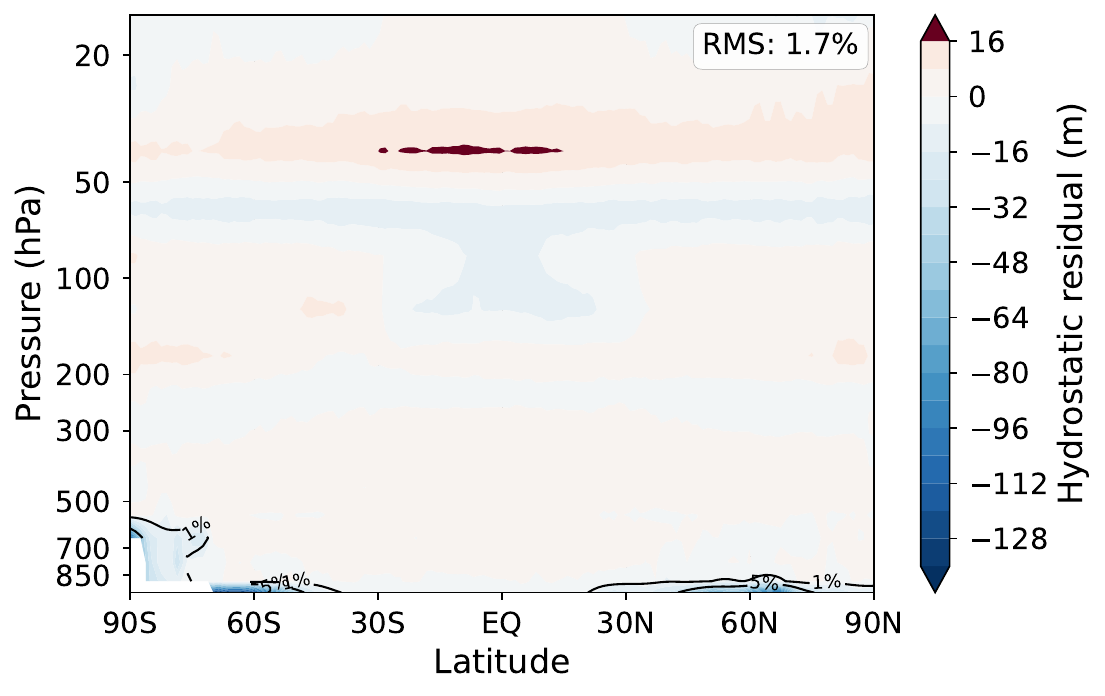} \caption{Hydrostatic balance residual, zonal and 85-year-rollout mean (computed locally per grid cell, then averaged), for $AC\text{-}SSP$, SSP5-3.4. Values closer to zero are better. The y-axis is pressure (surface at the bottom), log-scaled; the x-axis is latitude in degrees, with faint dotted reference lines at the standard 90S-90N latitudes. Black contour lines (lightly smoothed along latitude for legibility) mark where the residual reaches 1\%/5\% of the local hydrostatic layer thickness, and the boxed annotation reports the area-weighted root-mean-square (RMS) of that percentage across the whole column. The hydrostatic residual is small in both absolute and relative terms (area-weighted RMS $1.7$\% of layer thickness) and confined almost entirely to the surface and a narrow tropical band near $40$~hPa, indicating a largely physically consistent temperature-geopotential relationship elsewhere in the column.} \label{fig:hydrostatic}
\end{figure}

\begin{figure}
\includegraphics[width=1\textwidth]{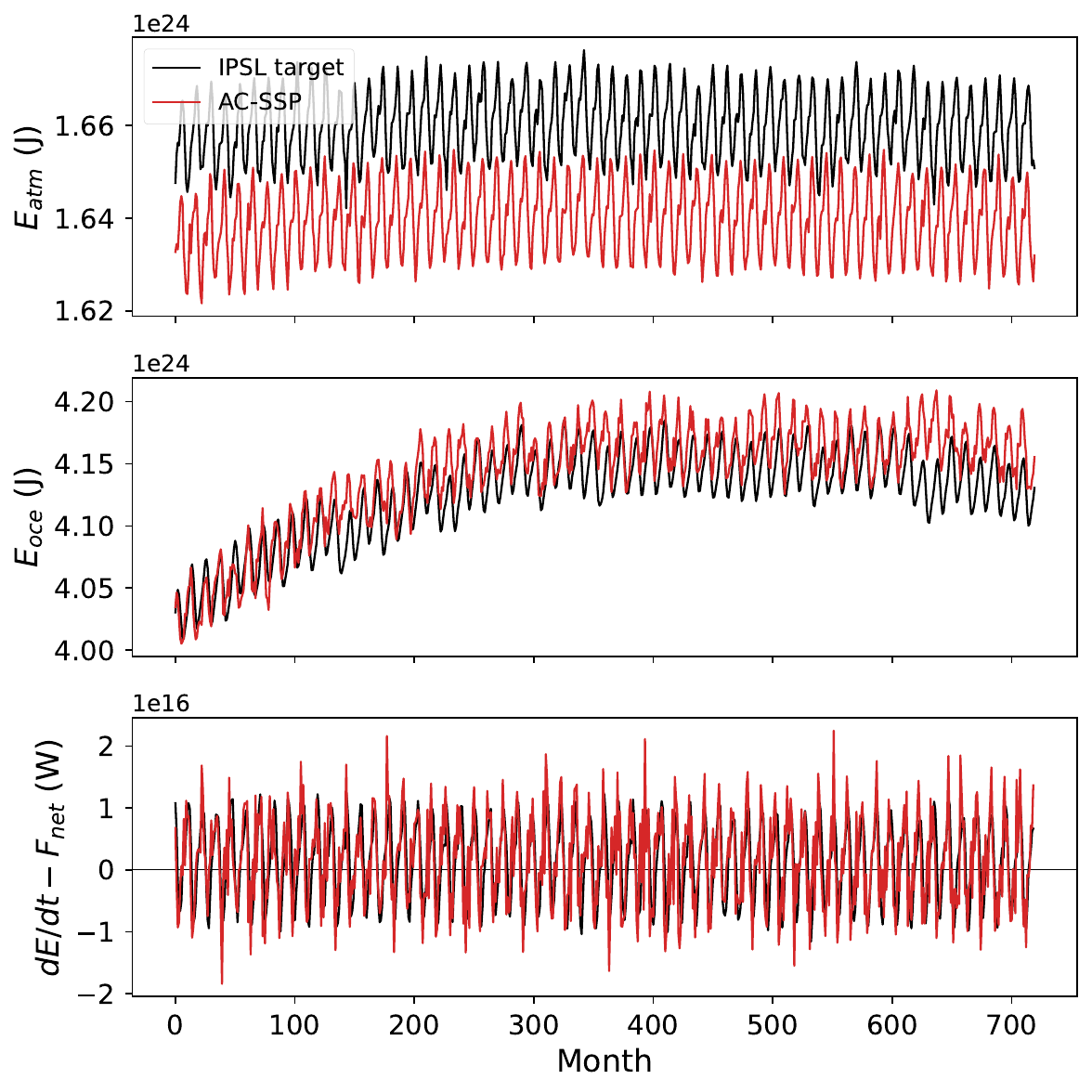} \caption{Coupled atmosphere--ocean energy budget for $AC\text{-}SSP$ (Energy budgets for SSP5-3.4: global $E_{\mathrm{atm}}$ and $E_{\mathrm{oce}}$ trajectories, model vs. target, and the $dE/dt - F_{\mathrm{net}}$ conservation residual. $E_{\mathrm{atm}}$ is the column-integrated atmospheric energy, the sum of sensible (internal), potential, kinetic, and latent heat contributions (\cref{eq:atm-energy}); $E_{\mathrm{oce}}$ is the upper-ocean heat content, the depth-integrated seawater temperature down to $147.4$~m (\cref{eq:ocean-energy}). $E_{\mathrm{atm}}$ and $E_{\mathrm{oce}}$ both closely track their targets with a consistent offset rather than diverging.} \label{fig:energy_budget_pf4_w050_80_10}
\end{figure}

\subsubsection{Physical Consistency}\label{sec:hydrostatic}

An autoregressive emulator trained on CMIP output is not constrained to obey the conservation laws of the system it emulates. We use two diagnostics to probe this for the ArchesClimate emulator: hydrostatic balance (the relationship between air temperature ($ta$) and geopotential height ($zg$)) and a coupled atmosphere--ocean energy budget. See \cref{app:metrics} for definitions of both diagnostics.

Results here are for a 5-member ensemble of $AC\text{-}SSP$, evaluated with the SSP 5-3.4 test scenario. As illustrated in \autoref{fig:hydrostatic}, a residual near zero indicates adherence to hydrostatic equilibrium. Rather than treating any nonzero residual as an outright violation of physical law, we treat this diagnostic as a measure of consistency with the climate model it is trained against. 

The vertical profile in \autoref{fig:hydrostatic} demonstrates that the hydrostatic residual is largely negligible, indicating a robust modeled relationship between temperature ($ta$) and geopotential ($zg$). A localized inconsistency appears near $40$~hPa in the Tropics, corresponding to sharp thermal gradients, though the absolute error remains low. More substantial residuals emerge at the surface, plausibly driven by a combination of unresolved geopotential gradients at the land-atmosphere boundary; attempts to mitigate this via logarithmic transformations of $zg$  yielded no measurable improvement in physical consistency (not shown). In the atmosphere--ocean energy budget (\autoref{fig:energy_budget_pf4_w050_80_10}), the atmospheric energy ($E_{\mathrm{atm}}$) closely tracks the target but exhibits a systematic offset. Decomposing the total $E_{\mathrm{atm}}$ mean absolute error ($2.74 \times 10^7$ J/m$^2$) clarifies this divergence: sensible heat contributes the majority ($\approx$58\%), followed by latent heat ($\approx$33\%) and potential energy ($\approx$16\%), while kinetic energy is negligible ($\approx$1.5\%). Please see \cref{app:energy_balance} for the calculations of these numbers. This error distribution is fundamentally rooted in a vertically non-uniform temperature bias: the model produces a tropospheric warm bias ($+0.2$ to $+0.6$~K) that reverses into a stratospheric cold bias ($-0.2$ to $-0.8$~K). Because $E_{\mathrm{atm}}$ is a mass-weighted column integral, the denser troposphere dictates the net positive error. Both diagnostics point to the observation that AC-SSP's physical inconsistencies are small, systematic biases tied to a specific, identifiable source (a vertically-structured temperature bias).

Finally, the ocean energy trajectory ($E_{\mathrm{oce}}$) exhibits slight drift and a model-target error of roughly similar magnitude to $E_{\mathrm{atm}}$. The instability in the value stems directly from the restricted depth of the emulated ocean ($147.4$~m is the deepest level represented; \autoref{sec:datasets}), which lacks the deep-ocean reservoir necessary to facilitate transient heat uptake \cite{roseEffectsOceanHeat2016}.

\subsection{Forcing Response: Physical Consistency of Single-Forcing Removal}\label{sec:forced_response}

We next check whether AC-SSP attributes the correct sign and magnitude of warming to each forcing, at the surface and across the atmospheric vertical profile. For each forcing (see \cref{sec:forcings} for a complete description), we hold that channel at its piControl (pre-industrial) value for the entire SSP5-3.4 rollout while leaving every other forcing untouched, and compare the resulting yearly global-mean $tas$ trajectory against the model's own all-forcings baseline. We hold forcings at piControl levels specifically to be consistent with DAMIP experiment design. We use the TCR procedure described in \cref{app:tcr_explanation} for CO$_2$, N$_2$O, and CH$_4$. For ozone and aerosols, we evaluate with explicit experiments, $hist-=piAer$ and $hist--stratO3$. Details for this procedure can be found in \cref{sec:ozone_aerosol_explanation}. 

\begin{figure}
\includegraphics[width=1\textwidth]{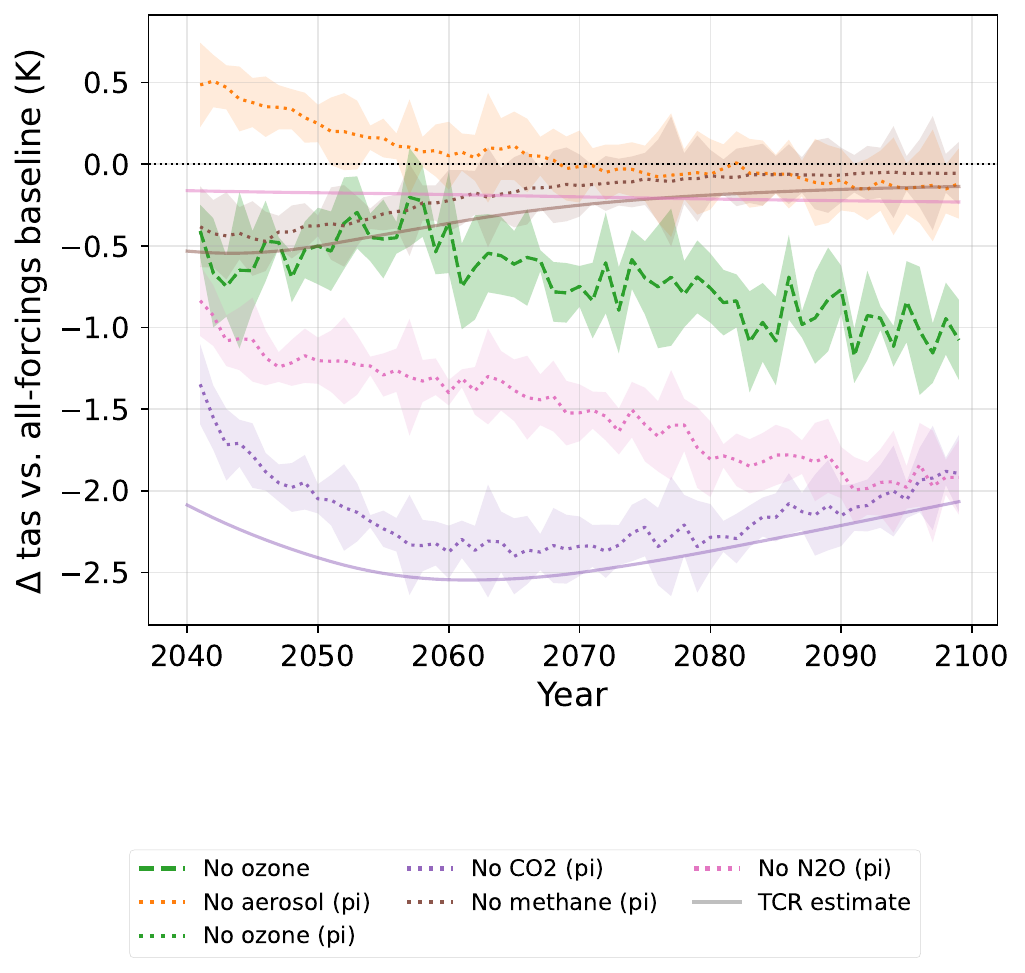}
\caption{Single-forcing piControl-held ablations under SSP5-3.4-over, $AC\text{-}SSP$: yearly global-mean $tas$ difference from the all-forcings baseline (dotted; mean across members, shaded band is $\pm1$ std across members). Each forcing is set to pre-industrial levels and held, the rest are left the same. For CO$_2$/methane/N$_2$O, the same-colored solid, semi-transparent line is that gas's physically expected magnitude from the ERF formula scaled by the TCR-derived sensitivity $\lambda_{\text{TCR}}$; no aerosol/ozone reference is shown, as the simple ERF formulas used here do not cover those forcings. Note the shaded band here is a single-year member spread and so includes each member's own interannual variability; it is not directly comparable in magnitude to \autoref{fig:whisker_534}'s decade-averaged band. AC-SSP recovers the correct sign of every forcing's contribution, but over-attributes warming to CO$_2$ and especially N$_2$O relative to their TCR references, while methane's response matches closely.} \label{fig:surface_forced_response}
\end{figure}

\autoref{fig:surface_forced_response} illustrates the forced response learned by AC-SSP. Holding CO$_2$ at piControl levels cools the trajectory by $2.00$~K by $2099$. This exceeds the $1.5$~K cooling predicted by the TCR reference, indicating that the model over-attributes aggregate warming to CO$_2$ specifically. The methane response, conversely, closely tracks the estimated TCR. The most severe discrepancy occurs with N$_2$O: withholding N$_2$O induces a $2.0$~K cooling by $2100$, overshooting the $0.2$~K TCR estimate by a factor of ten.

We attribute this CO$_2$ sensitivity to severe collinearity between the CO$_2$ and N$_2$O forcing trajectories within the training corpus. Because CO$_2$ and N$_2$O concentrations rise synchronously across the SSP scenarios, the DAMIP historical ensemble primarily constrains the aggregate greenhouse gas response rather than isolating individual constituent contributions. While the idealized \textit{1pctCO2} experiment theoretically anchors the isolated CO$_2$ signal, it provides no equivalent independent constraint for N$_2$O. Consequently, the model fails to establish an independent physical sensitivity for N$_2$O, defaulting instead to a confounded correlation. Resolving this degeneracy and establishing a robust N$_2$O representation would necessitate expanding the training dataset to include explicit single-forcing N$_2$O ablations.

\begin{figure}
\includegraphics[width=1\textwidth]{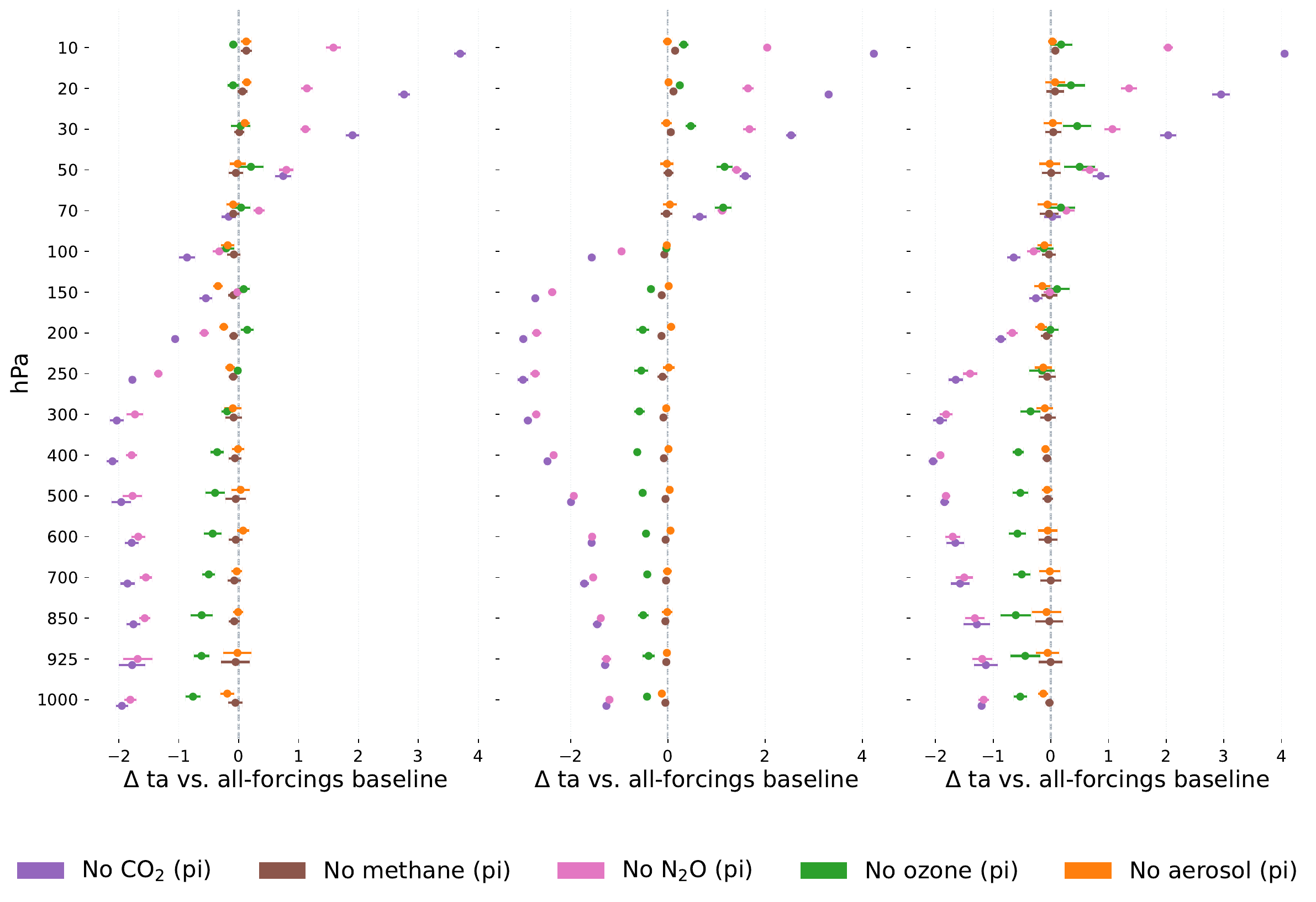}
\caption{Atmospheric-column response to each single-forcing piControl-held ablation, $AC\text{-}SSP$, SSP5-3.4 averaged over the last decade (2090-2100): area-weighted mean $ta$ difference from the model's own all-forcings baseline, by pressure level, left to right North (18.3°N-90°N) / Tropics (19.5°S-18.3°N) / South (90°S-19.5°S) latitude bands. Shaded bands show $\pm1$ std across ensemble members, computed after averaging each member over its own trailing decade. This suppresses independent year-to-year noise relative to \autoref{fig:surface_forced_response}'s single-year band. Removing CO$_2$ or N$_2$O produces the same physically-expected sign reversal (tropospheric cooling, stratospheric warming) at every latitude band, while methane and aerosol produce comparatively little vertical-column response.} \label{fig:whisker_534}
\end{figure}

The same piControl-held ablations, resolved across the atmospheric column rather than collapsed to a global-mean surface time series, are shown in \autoref{fig:whisker_534}, for $AC\text{-}SSP$ for SSP5-3.4. Removing CO$_2$ or N$_2$O produces the largest and most vertically coherent response of the five forcings, and both share the same characteristic shape: tropospheric cooling (roughly $-1.2$ to $-2.0$~K at $1000-500$~hPa) that reverses sign through the upper atmosphere into a pronounced stratospheric warming ($+0.7$ to $+0.8$~K at $50$~hPa, rising to $+1.6$ to $+4.2$~K at $10$~hPa). This sign reversal is physically expected: removing a well-mixed greenhouse gas removes both its tropospheric warming effect and its stratospheric cooling effect, so the piControl-held column develops a warmer stratosphere alongside a cooler troposphere. CO$_2$'s effect is somewhat larger than N$_2$O's at most levels and regions, consistent with its larger radiative forcing contribution over this period. Methane, by contrast, produces a negligible response at every level and region, despite showing a more substantial surface response in \autoref{fig:surface_forced_response}. 

Ozone exhibits a modest tropospheric cooling effect (-0.4 to -0.8 K at 1000 hPa). By the end of the century, it will drive multiple interacting feedbacks with divergent climate impacts; for instance, the removal of ozone in the tropical upper atmosphere is projected to induce localized cooling. Regarding aerosols, our current representation is restricted to a two-dimensional atmospheric load rather than a vertically explicit column. However, because the aerosol forcing signal under the SSP5-3.4 scenario remains minimal for the 2090–2100 period, the deviation from pre-industrial control (piControl) levels is expected to be negligible.

\begin{figure}
\includegraphics[width=1\textwidth]{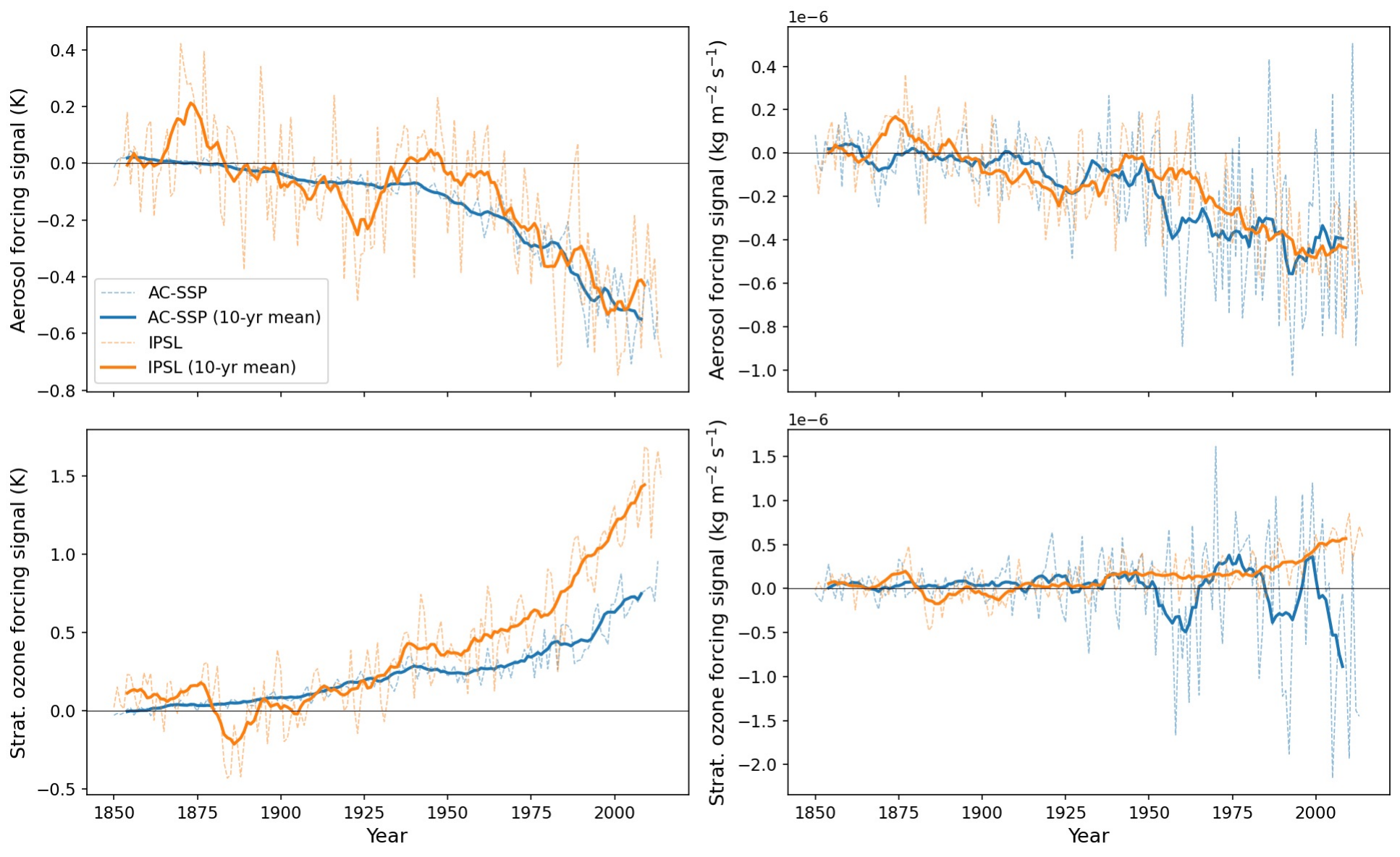}
\caption{$AC\text{-}SSP$ vs.  IPSL, $1850$-$2014$: (top row) aerosol forcing signal, generated \textit{historical} minus generated $hist--piAer$ global-mean value, against the same difference computed from the real IPSL data; (bottom row) stratospheric-ozone forcing signal, analogous but for $hist--stratO3$ (an additive DAMIP construction, so it reflects the combined effect of every non-strat-ozone forcing, not stratospheric ozone alone -- see \autoref{sec:forced_response}). Left column: $tas$ (K); right column: $pr$ (kg m$^{-2}$ s$^{-1}$). Because both the real, single-realization IPSL series and AC-SSP's own generated series are noisy year to year at this global-mean resolution, each panel shows a $10$-year centered rolling mean of each series (solid) alongside the corresponding raw one (faint dashed). Despite never training on either DAMIP experiment, AC-SSP's smoothed trend reproduces the sign, magnitude, and long-term trend of both real $tas$ forcing signals, tracking the smoothed IPSL trend closely; the $pr$ signal is far noisier for both series and AC-SSP's precipitation response is less cleanly separated from noise than its temperature response.}
\label{fig:aerosol_strato3_forcing}
\end{figure}

To provide analysis of the forced response of ozone and aerosols in AC-SSP, we test against two physically consistent single-forcing trajectories, rather than a piControl-held ablation of an all-forcings baseline as performed above. A complete description of this experiment is provided in Section \ref{sec:ozone_aerosol_explanation}. As illustrated in \cref{fig:aerosol_strato3_forcing}, the aerosol forcing experiment generated by AC-SSP closely tracks the smoothed mean trend of the IPSL and scales appropriately throughout the historical period. Conversely, the modeled ozone response captures the trend effectively until the latter portion of the simulation, where the IPSL signal becomes stronger than the AC-SSP. This divergence likely stems from an incomplete representation of stratospheric ozone, as our configuration incorporates only a subset of available forcings (10 out of the 66 layers, evenly distrbiuted throughout the atmospheric column, are included as forcings). Nevertheless, capturing the positive trend is an encouraging indication that the atmospheric column is adequately represented, given that stratospheric dynamics successfully propagate down to the surface. Precipitation exhibits a comparable trajectory, with similar deterioration at the end of the stratospheric ozone experiment. 

Interestingly, internal variability is highly pronounced from the onset of each IPSL experiment, presumably reflecting an initial spin-up period. Our emulator requires substantial time to properly generate sufficient internal variability between the two runs, as they are initialized with the same noise token and only differ in the forcing held to pre-industrial levels. While both states initialize with marginal discrepancies, these differences amplify over time, becoming particularly pronounced in the precipitation fields. Implementing a dedicated spin-up period for the emulator could address this early variance. Alternatively, a more engineered solution would involve applying a distinct noise-conditioning token between the two experiments.

\subsection{Out-of-Distribution Validation}\label{sec:585}

This section evaluates out-of-distribution generalization from two angles: SSP5-8.5, a smooth shift in CO$_2$ forcing beyond anything seen in training, and the idealized abrupt-4xCO2 experiment, a discontinuous forcing step. The first one probes whether the model extrapolates along a familiar direction pushed further than it has seen, the other probes whether it can respond correctly to a forcing discontinuity unlike anything in its training trajectories.

As the most extreme high-emission pathway in the CMIP6 suite, SSP5-8.5 represents a climatic regime characterized by high CO$_2$ concentrations that provide a high climate change signal-to-noise ratio \cite{deutloffHighProbabilityTriggering2025}. Here, we use this experiment to assess how specific forcing agents influence model extrapolation when greenhouse gas forcing conditions exceed value ranges seen during training. The following analysis focuses on density plots generated for the $2090$–$2100$ period, as well as rollouts for the entire period.

\begin{figure}
\includegraphics[width=1\textwidth]{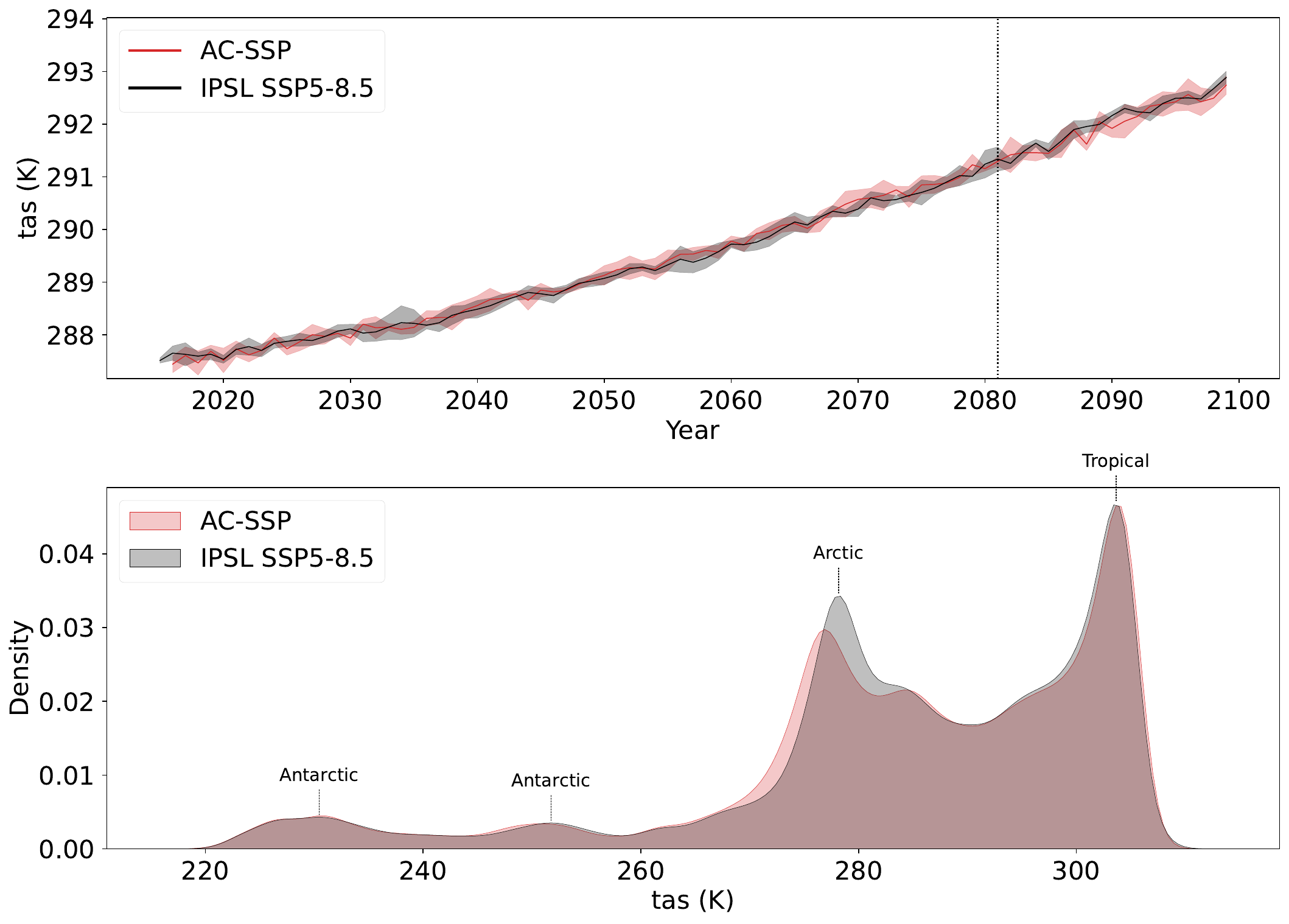} \caption{Out-of-distribution performance under the SSP5-8.5 high-emission scenario, $AC\text{-}SSP$ (AC-SSP vs.\ the IPSL target, a 4-member model ensemble vs.\ the 4-member IPSL ensemble. (Top) Time series of global-mean yearly $tas$ (lat-weighted spatial average over all grid cells, then a mean $\pm$ std band across members/realizations) from $2015$-$2100$; the dotted vertical line marks $2081$, the first year SSP5-8.5's CO$_2$ concentration exceeds the maximum CO$_2$ level reached anywhere in the training data (SSP3-7.0's own maximum, reached at its final year, 2100). (Bottom) Probability density functions (PDFs) of grid-cell yearly-mean $tas$: computed by pooling every grid cell, member/realization, and year of the final decade into a single distribution, with each of the four modes labeled by the temperature regime it corresponds to (Antarctic interior, Antarctic coastal/peripheral, Arctic, Tropical). AC-SSP tracks the IPSL target's warming trend and grid-cell temperature distribution closely across the whole 85-year rollout and all four modes, with the Arctic mode's density being the only shortfall.} \label{fig:extrapolation}
\end{figure}

The density plots for the stable base models, shown in the lower panel of \autoref{fig:extrapolation}, show a multimodal distribution of surface temperature at each grid cell. We verified each mode's identity directly, by checking which grid cells (and their latitudes) actually populate each mode's temperature window. The four modes are: the Antarctic interior (coldest, elevation-driven), Antarctic's coastal/peripheral regions (second-coldest), the Arctic (warmer than either Antarctic mode, since it is ocean/ice-moderated with no comparable elevation effect), and the tropics (warmest); mid-latitudes populate the smooth area between the Arctic and Tropic modes. Of the four modes, the Arctic mode is the least accurate, and we attribute this to a combination of factors rather than a single cause. First, the extreme forcing of the SSP5-8.5 pathway pushes the Arctic furthest outside the model's training distribution. Because the region warms disproportionately faster than the global mean due to Arctic amplification \cite{previdiArcticAmplificationClimate2021}, end-of-century grid-cell $tas$ values reach magnitudes entirely absent from the historical and moderate SSP training sets (\autoref{sec:datasets}). Second, the polar regions also carry substantially higher year-to-year variability than the tropics or mid-latitudes, because of sea-ice/open-ocean transitions. Thirdly, this effect is further increased due to the fact that a given grid cell spans less physical distance at high latitudes, so the same fixed-resolution grid resolves comparatively finer, noisier spatial structure there; a model trained with a loss that is latitude-aware will tend to under-weight this high-variance, small-area region relative to the tropics, where each grid cell covers far more physical area and dominates a weighted loss. We currently apply latitude-area weighting in the training loss, which we expect compounds the Arctic mode's underrepresentation.

\subsubsection{Abrupt-4xCO2} 

As a complementary stress test, we evaluate the models using the idealized abrupt-4xCO2 experiment. Here, atmospheric CO$_2$ is instantaneously quadrupled from its $1850$ pre-industrial baseline and held constant, confronting the model with a forcing discontinuity rather than the continuous, physical ramps present in the training data (\autoref{sec:datasets}). We assess the global-mean near-surface air temperature ($tas$) trajectory against the IPSL target, holding all non-CO$_2$ forcings at $1850$ levels. To determine whether the energy-score objective itself degrades the transient response, characterized by rapid initial warming followed by a slow, ocean-mediated equilibration, we compare AC-SSP against a deterministic baseline ($pf=4$, MSE loss, single ensemble member).

\begin{figure}\includegraphics[width=1\textwidth]{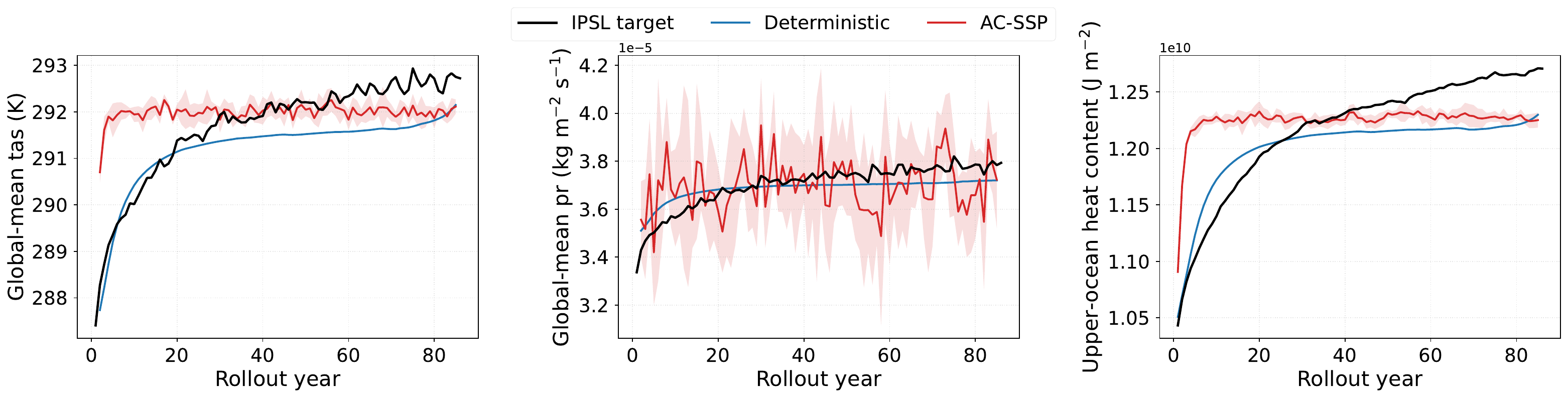}\caption{Abrupt-4xCO2, global-mean $tas$ (left), $pr$ (middle), and upper-ocean heat content (right, depth-integrated as in \cref{eq:ocean-energy}) over an 80-year rollout, against the IPSL target (r2i1p1f1, black): the deterministic model (with all other parameters held constant) versus AC-SSP. The deterministic model reproduces the target's slow, ocean-mediated transient warming in all three quantities, while AC-SSP instead jumps almost immediately to a near-final state and plateaus.} \label{fig:abrupt4xco2_det_vs_es}\end{figure}

The deterministic baseline captures the slow signal: global-mean $tas$ rises smoothly by $3.50$~K (from a mean of $288.28$~K during years $1$-$5$; to a mean of $291.78$~K during years $76$-$80$), mirroring the target's trajectory despite underestimating the final magnitude. AC-SSP, conversely, effectively bypasses the transient phase entirely. It reaches within $1$~K of its final state in the first five years and plateaus, warming by only $0.98$~K over the remainder of the period. Global-mean precipitation ($pr$) exhibits an identical failure mode, with AC-SSP capturing only half of the deterministic model's smooth rise alongside exacerbated inter-annual noise (\autoref{fig:abrupt4xco2_det_vs_es}, middle panel). Upper-ocean heat content (\autoref{fig:abrupt4xco2_det_vs_es}, right panel; computed as in \cref{app:energy_balance}) shows the same phenomenon. This indicates simply adding ocean variables to a model does not help it learn slower-than-atmosphere temporal dynamics. 

AC-SSP does not asymptotically approach an equilibrium; it immediately jumps to its final state. Because both models utilize identical architectures, this discrepancy stems from the probabilistic objective. We hypothesize that the energy score, by incentivizing a state distribution conditioned strictly on the concurrent forcing, biases the model toward a forcing-conditioned equilibrium rather than a time-lagged trajectory. The deterministic model, unconstrained by distributional requirements, preserves the lagged trajectory learned during push-forward finetuning.

If this is correct, fixing this requires addressing the energy-score loss. We propose three untested but physically motivated directions: modeling the energy score as a residual atop a frozen deterministic base trajectory, scoring the push-forward rollout over the full temporal trajectory rather than frame-by-frame to penalize implausible paths (i.e., paths that quickly jump to some equilibrium) or conditioning the model on an extended historical state window to provide explicit rate-of-change information.

\subsection{CanESM5 application} \label{sec:canesm5_application}

To assess whether AC-SSP generalizes beyond the IPSL-CM6A-LR dataset, we adapt the preprocessing pipeline for CanESM5 \cite{swartCanadianEarthSystem2019a}, an alternative CMIP6 model. This model was selected because it had the dataset most similar to IPSL among available CMIP models. This adaptation requires several structural changes prior to model training. The level grid, the ocean vertical grid, the surface variable set, and the available experimental protocols are all adjusted to fit the CanESM5. Please refer to \cref{app:canesm5_ipsl_members} for details on the data and preprocessing of the CanESM5 model.

\begin{figure}[h!]
\includegraphics[width=0.8\textwidth]{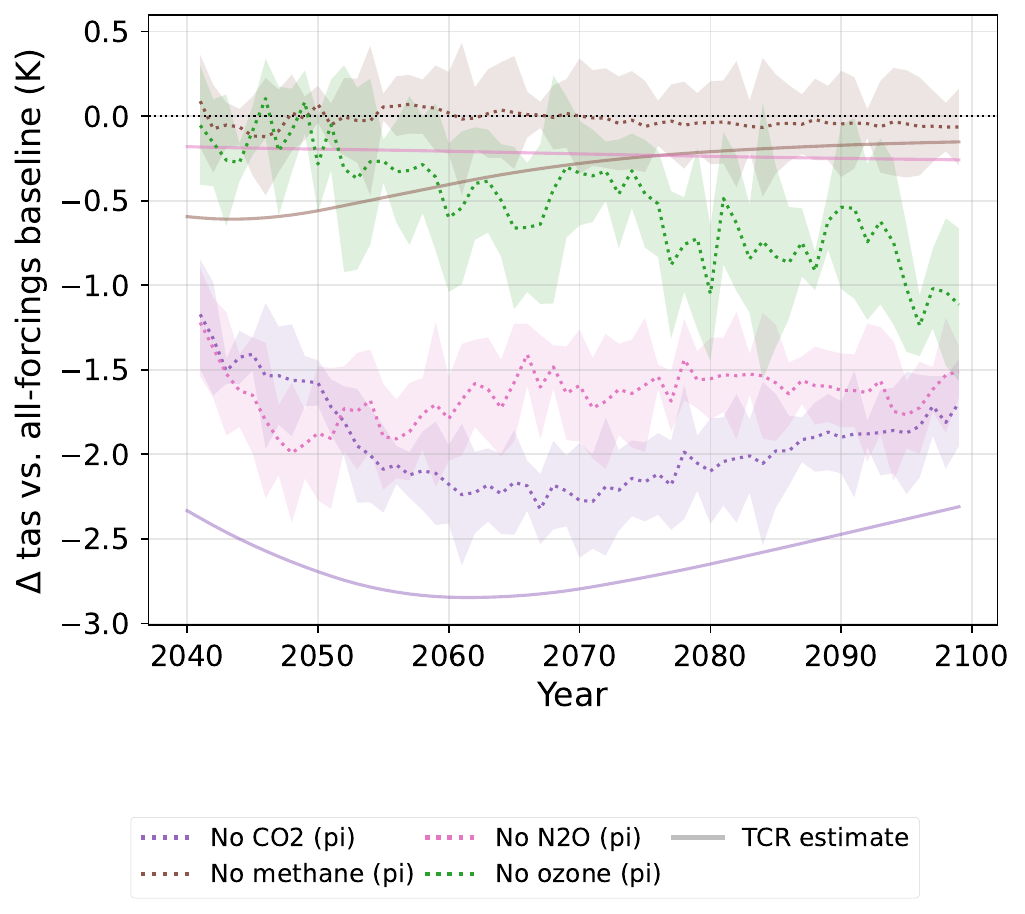} \caption{CanESM5 energy-score model, single-forcing piControl-held ablations under SSP5-3.4: yearly global-mean $tas$ difference from the all-forcings baseline. Each forcing is set to pi-control levels and held for the entire rollout, the rest are left the same. CanESM5's aerosol forcing is excluded from training entirely (see above), so no aerosol ablation exists here. For CO$_2$/CH$_4$/N$_2$O, the same-colored solid, semi-transparent line is that gas's physically expected magnitude from the ERF formula scaled by CanESM5's own TCR-derived sensitivity $\lambda_{\text{TCR}}$ ($\mathrm{TCR}=2.74$K, \cref{app:tcr_explanation}); no comparable reference exists for ozone (\cref{app:tcr_explanation}). CanESM5's single-forcing attribution behavior mirrors IPSL's, indicating this behavior is a property of the training data and objective, not an IPSL-specific feature.} \label{fig:canesm5_forced_response}
\end{figure}

\autoref{fig:canesm5_forced_response} illustrates the single-forcing attribution test applied to the CanESM5 energy-score model under the SSP5-3.4 scenario. The results indicate that despite architectural differences in the base model, the learned dynamics remain consistent with the IPSL baseline: holding other forcings constant results in relative cooling, and the CO$_2$, CH$_4$, and N$_2$O responses each track their own ERF$\times$TCR reference (using CanESM5's own $2.74$K TCR, \cref{app:tcr_explanation}) with the same qualitative over-attribution to CO$_2$/N$_2$O and close methane agreement seen for IPSL. Ozone has no comparable quantitative reference (\cref{app:tcr_explanation}), so that trace remains only qualitatively comparable to the IPSL results. Holding ozone at pre-industrial levels yields cooling trajectories similar to those observed in the IPSL model. Ultimately, these observations suggest that the framework captures comparable underlying dynamics even when applied to an ESM with differing physical parameterizations.

\begin{figure}[h!]
\includegraphics[width=1\textwidth]{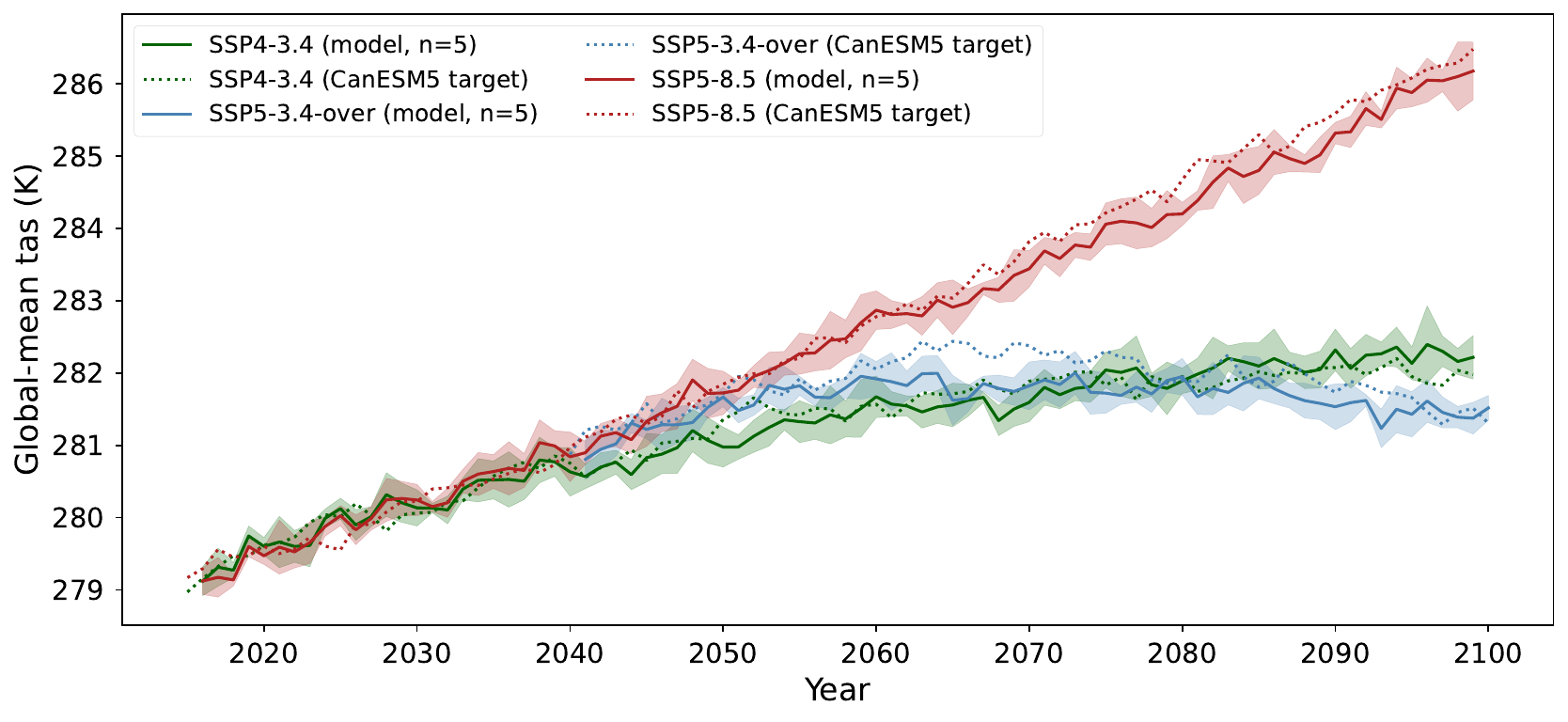} \caption{CanESM5 energy-score model, yearly global-mean $tas$ (mean $\pm$ std band across ensemble members where applicable) for SSP4-3.4, SSP5-3.4, and SSP5-8.5 rollouts, each plotted against its own CanESM5 target realization. AC-SSP tracks all three CanESM5 scenarios closely, including the SSP5-8.5 pathway, showing the framework's generalization ability is not specific to IPSL-CM6A-LR.} \label{fig:canesm5_multiscenario}
\end{figure}

\autoref{fig:canesm5_multiscenario} shows the performance of the CanESM5 deterministic energy-score model across three distinct evaluation scenarios: SSP4-3.4 (in-distribution validation), SSP5-3.4-over (an out-of-distribution overshoot pathway absent from the training set), and SSP5-8.5 (the extrapolation stress test previously applied to the IPSL model in \autoref{sec:585}). Consistent with the IPSL results, the CanESM5 model tracks the target data reasonably well across the evaluated scenarios. While the temperature peak within the SSP5-3.4-over pathway is marginally underestimated, the broader trends are captured with sufficient accuracy.

\section{Discussion and Conclusion}

We present AC-SSP, a deep-learning climate emulator capable of separate forced responses from internal climate variability while maintaining explicit spatial dynamics. In contrast to existing baselines such as MESMER-M, AC-SSP resolves the full three-dimensional atmospheric profile. Furthermore, its generative capacity enables the production of diverse ensemble members from single initializations (e.g., IPSL trajectories), thus facilitating rigorous probabilistic analyses of deterministic scenarios. 

We capture reasonable global trends for several SSP scenarios not seen at training time. We also show that the model has a reasonable forced response for the forcings included, although the impact of N$_2$O is overestimated. Consequently, decorrelating variables that covary in the historical record is essential for robust generalization. This is reinforced by \cite{clarkDisentanglingEffectsSea2026a}, who demonstrated that models trained on explicitly decorrelated data generalize far better to counterfactual scenarios. Expanding our training mixture with idealized point-source experiments—such as VolMIP \cite{zanchettinModelIntercomparisonProject2016} for stratospheric aerosol injections or ISA-MIP \cite{timmreckInteractiveStratosphericAerosol2018} for aerosol-climate interactions would further disentangle these signals. This could also be addressed through causal methods that could better learn the impact of the forcings on climate \cite{hickmanCausalClimateEmulation2025a}.

Because our training data intrinsically couples climate outputs with their generative forcings, we hypothesize that these prescribed forcings function primarily as attention-guiding mechanisms rather than wholly independent information sources. This is underlined by the fact that a model without forcings is numerically stable but does not follow any coherent path. This also supports the idea that our generative backbone can model internal variability without the support of forcings. 

During these experiments, we saw a persitent bias in the Arctic region. To fix this, it would plausibly require several changes beyond the scope of the current model: an area-weighted loss term to stop the tropics from dominating gradient updates; a native icosahedral or HEALPix grid representation, which does not artificially inflate polar spatial resolution and avoids the pole singularity altogether and using rotary positional embeddings (RoPE) in place of the backbone's current positional encoding, which may generalize better to the extreme positions the polar regime effectively represents \cite{suRoFormerEnhancedTransformer2023}. 

When evaluating model robustness, we distinguish two distinct forms of out-of-distribution behavior. As demonstrated in \autoref{sec:585}, AC-SSP extrapolates competently to the global temperature trend of the SSP5-8.5 scenario, where the global CO$_2$ trajectory persistently exceeds the training domain. The other is the large initial increase of CO$_2$ seen with the abrupt4xCO2. This illuminates interesting questions about the role of the ocean in this model. The ocean, in nature, acts as a buffer and regulates and slows temperature \cite{roseEffectsOceanHeat2016}. We do not see any stopping forces in this model, as the abrupt4xCO2 generation quickly snaps to the expected equilibrium. It therefore may be necessary to include physical constraints, or include a more comprehensive representation of the ocean. 

Ultimately, understanding irreducible scenario uncertainty is crucial for effective climate adaptation. By demonstrating that integrating forcings into a generative framework accurately captures shifting atmospheric dynamics, including complex, policy-relevant overshoot trajectories such as SSP5-3.4, AC-SSP provides a critical tool for exploring diverse human pathways. These findings establish a methodological foundation for developing computationally efficient, probabilistically robust emulators for long-term climate projection, thereby enabling the rapid scenario generation necessary for policymakers to assess how immediate mitigation strategies alter centennial climate outcomes.

\section*{References}

\appendix

\section{Appendix}

\subsection{Climate Model Data}\label{sec:datasets}

\begin{longtable}{p{0.14\textwidth} p{0.55\textwidth} p{0.08\textwidth} p{0.15\textwidth}}
\caption{Overview of IPSL scenario runs considered in the training and testing regime of AC-SSP. The scenario name and description identify the type of experiment run by IPSL; AC-SSP trains and tests on the outputs of a subset of those runs. All SSP runs are part of the ScenarioMIP experiments. ``Ens. Mem.'' stands for the number of ensemble members, i.e., repeated runs of the same scenario experiment, available. All runs provide monthly data for the years indicated. SSP4-3.4 is held out for validation, Abrupt-4xCO2 and SSP5-8.5 are held out for OOD validation, SSP5-3.4 is held out for testing, while all other scenarios and their ensembles are used for training.}\label{table:scenarios}\\
\toprule
Scenario Name & Description & Ens. Mem. & Years \\
\midrule
\endfirsthead
\multicolumn{4}{c}{\tablename\ \thetable{} -- continued from previous page}\\
\toprule
Scenario Name & Description & Ens. Mem. & Years \\
\midrule
\endhead
\midrule
\multicolumn{4}{r}{\textit{Continued on next page}}\\
\endfoot
\bottomrule
\endlastfoot

\textit{piControl} & Pre-industrial control simulation, used as a baseline with constant pre-industrial conditions & 1 & 1850, repeated for 150~years\\

\textit{historical} &  Simulation of the climate using observed historical forcings& 10 &1850-2014\\

\textit{SSP1-1.9} & A scenario where the world shifts toward sustainable development and net-zero emissions, successfully limiting global warming to approximately 1.5°C by 2100 & 4 & 2015-2100 \\

\textit{SSP1-2.6} & Future scenario following a low-emissions pathway (SSP1-2.6), aiming for strong mitigation. & 5 &2015-2100 \\

\textit{SSP2-4.5} & Intermediate emissions scenario (SSP2-4.5), representing moderate mitigation and socioeconomic development & 6 &2015-2100 \\

\textit{SSP3-7.0} & High emissions scenario (SSP3-7.0), reflecting regional rivalry and limited mitigation & 10 &2015-2100\\

\textit{SSP4-3.4} & Medium-to-high emissions scenario (SSP4-3.4), focusing on high inequality and partial mitigation. & 1 &2015-2100\\

\textit{SSP4-6.0} & A scenario characterized by high socioeconomic inequality and regional stratification, resulting in moderate climate policy success and a warming of roughly 2.7°C by 2100. &6 & 2015-2100\\

\textit{SSP5-3.4} &  SSP5-3.4 (Overshoot): A fossil-fuel-driven scenario that initially follows a high-emission path (SSP5-8.5) until 2040, followed by an aggressive, late-century shift to carbon removal to bring forcing down to 3.4 $W/m^2$& 1 &2015-2100\\

\textit{SSP5-8.5} & Very high emissions scenario (SSP5-8.5), representing a “business-as-usual” pathway with minimal mitigation & 4 & 2015-2100 \\

\textit{Abrupt-4xCO2} &  A idealized scenario that begins with a 4x step of CO$_2$ concentrations, starting from initial conditions at 1850 & 1 & 1850, repeated for 100~years \\

\end{longtable}

\begin{longtable}{l l p{8cm}}
\caption{Variables used in AC-SSP. Variables are categorized by Earth system component: Surface, Ocean, and Atmosphere. Each entry includes the CMIP-standardized name and a corresponding description. For oceanic and atmospheric components, the model incorporates all defined vertical depths and levels, respectively. All listed variables are predicted autoregressively.}\label{table:variables}\\
\toprule
\textbf{Category} & \textbf{Variable} & \textbf{Long Name} \\
\midrule
\endfirsthead
\multicolumn{3}{c}{\tablename\ \thetable{} -- continued from previous page}\\
\toprule
\textbf{Category} & \textbf{Variable} & \textbf{Long Name} \\
\midrule
\endhead
\midrule
\multicolumn{3}{r}{\textit{Continued on next page}}\\
\endfoot
\bottomrule
\endlastfoot

\multirow{8}{*}{\textbf{Surface}}
 & \textit{vas} & Northward Near-Surface Wind \\
 & \textit{uas} & Eastward Near-Surface Wind \\
 & \textit{pr} & Precipitation \\
 & \textit{ps} & Surface air pressure \\
 & \textit{tas} & Near-surface air temperature \\
 & \textit{evspsbl} & Evaporation including sublimation and transpiration \\
 & \textit{net\_surface\_flux} & Total positive downward flux between ocean and atmosphere \\
 & \textit{psl} & Sea level pressure \\

\midrule

\multirow{2}{*}{\textbf{Ocean}} & \multicolumn{2}{l}{\textit{Depths (m): 0.51, 3.86, 8.09, 13.99, 22.76, 35.74, 53.85, 77.61, 108.03, 147.41}} \\
\cmidrule(lr){2-3}
 & \textit{thetao} & Ocean temperature \\

\midrule

\multirow{7}{*}{\textbf{Atmospheric}} & \multicolumn{2}{p{10cm}}{\textit{Levels (hPa): 1000, 925, 850, 700, 600, 500, 400, 300, 250, 200, 150, 100, 70, 50, 30, 20, 10}} \\
\cmidrule(lr){2-3}
 & \textit{hus} & Specific humidity \\
 & \textit{ta} & Air temperature \\
 & \textit{ua} & Eastward wind \\
 & \textit{va} & Northward wind \\
 & \textit{zg} & Geopotential height \\

\end{longtable}

\subsection{Forcing Dropout Logic Example}\label{app:forcing_dropout_ex}

Here we provide a concrete example of the Forcing Dropout logic used in AC-SSP, introduced in \cref{sec:architecture}. On the remaining 1-$p_{\text{all}}$ samples, we drop the forcings with a probability of $p_{\text{drop}}$. The two parameters are therefore only meaningful jointly: at $p_{\text{all}}=1$, $p_{\text{drop}}$ has no effect at all, since the masking branch is never reached; at $p_{\text{all}}=0$, the model's forcing exposure is controlled by $p_{\text{drop}}$. This allows us to ensure with some explicit probability how much the model sees the full forcing ($p_{\text{all}}$). For example, if we have $p_{\text{all}}=0.8$ and $p_{\text{drop}}=0.2$, then the model will see all of the forcings 80\% of the time; on the remaining 20\% of steps, each forcing independently has a 20\% chance of being dropped: P(drop) = $(1-p_{\text{all}})\cdot p_{\text{drop}} = 0.2 \times 0.2 = 0.04$. With these parameters, any given forcing is dropped 4\% of the time during training.

\subsection{Stability Across Random Seeds (Energy-Score Model)}\label{app:seed_stability_energy_score}

We check for the energy-score used throughout this paper: three otherwise-identical training runs of AC-SSP differing only in training-time random seed, all at step 22000, evaluated as 5-member ensembles on SSP4-3.4.

\begin{table}[ht]
\centering
\begin{tabular}{llcccc}
\toprule
\textbf{Scenario} & \textbf{Model} & \textbf{RMSE (K)} & \textbf{\shortstack{Spatial Std.\\\\ Deviation}} & \textbf{\shortstack{Ensemble\\\\ Spread}} & \textbf{\shortstack{Inter-annual\\\\ Variability}} \\
\midrule
\multirow{5}{*}{\textbf{SSP4-3.4}}
    & seed1 & 1.0356 & 13.4601 & 0.4600 & 0.6160 \\
    & seed2 & 1.1871 & 13.4339 & 0.5373 & 0.6283 \\
    & seed3 & 1.2591 & 13.6251 & 0.5872 & 0.6355 \\
    & $MESMER\text{-}M$ & 0.9390 & 13.2717 & 0.1179 & 0.7685 \\
    & IPSL & n/a & 13.2520 & n/a & 0.6337 \\
\bottomrule
\end{tabular}
\caption{Per-metric comparison of across three training seeds, step-22000, 5-member ensemble each, SSP4-3.4.}
\label{tab:seed_stability_energy_score}
\end{table}

This energy-score model does not converge tightly across seeds. RMSE ranges from $1.0356$K (\texttt{seed1}) to $1.2591$K (\texttt{seed3}), a $22\%$ relative spread between the best and worst seed. We do not have an explanation for why the three seeds order themselves so consistently across all three metrics rather than showing the seed-independent scatter we would expect from pure initialization noise. One possibility is that the energy-score loss's added stochasticity (\autoref{sec:energy_score_loss}) produces seed-unstable training dynamics. We note this as a limitation of the results reported for this model throughout this paper:.

\subsection{Forcing All-Present Probability at the Extremes: $p_{\text{all}}=0.8$ vs.\ $1.0$}\label{app:ap_08_vs_10_forced_response}

\begin{figure}[ht]
\centering
\includegraphics[width=1\textwidth]{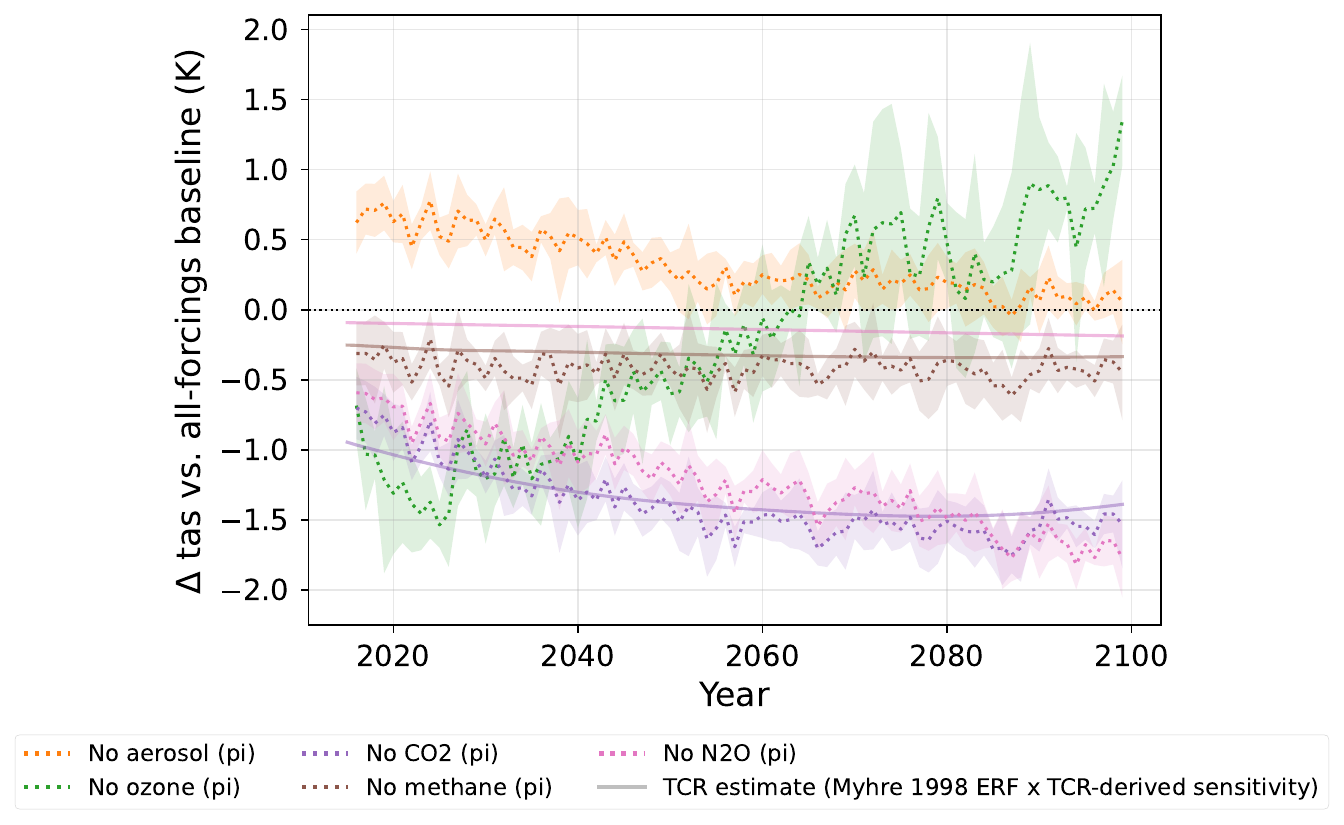}
\caption{$p_{\text{all}}=0.8$ (Single-forcing piControl-held ablations under SSP4-3.4, same methodology as \autoref{fig:surface_forced_response}. At $p_{\text{all}}=0.8$, methane's, ozone's and aerosol's ablation trace has the correct sign and tracks its TCR reference closely.}
\label{fig:w050_80_10_434_forced_response}
\end{figure}

\begin{figure}[ht]
\centering
\includegraphics[width=1\textwidth]{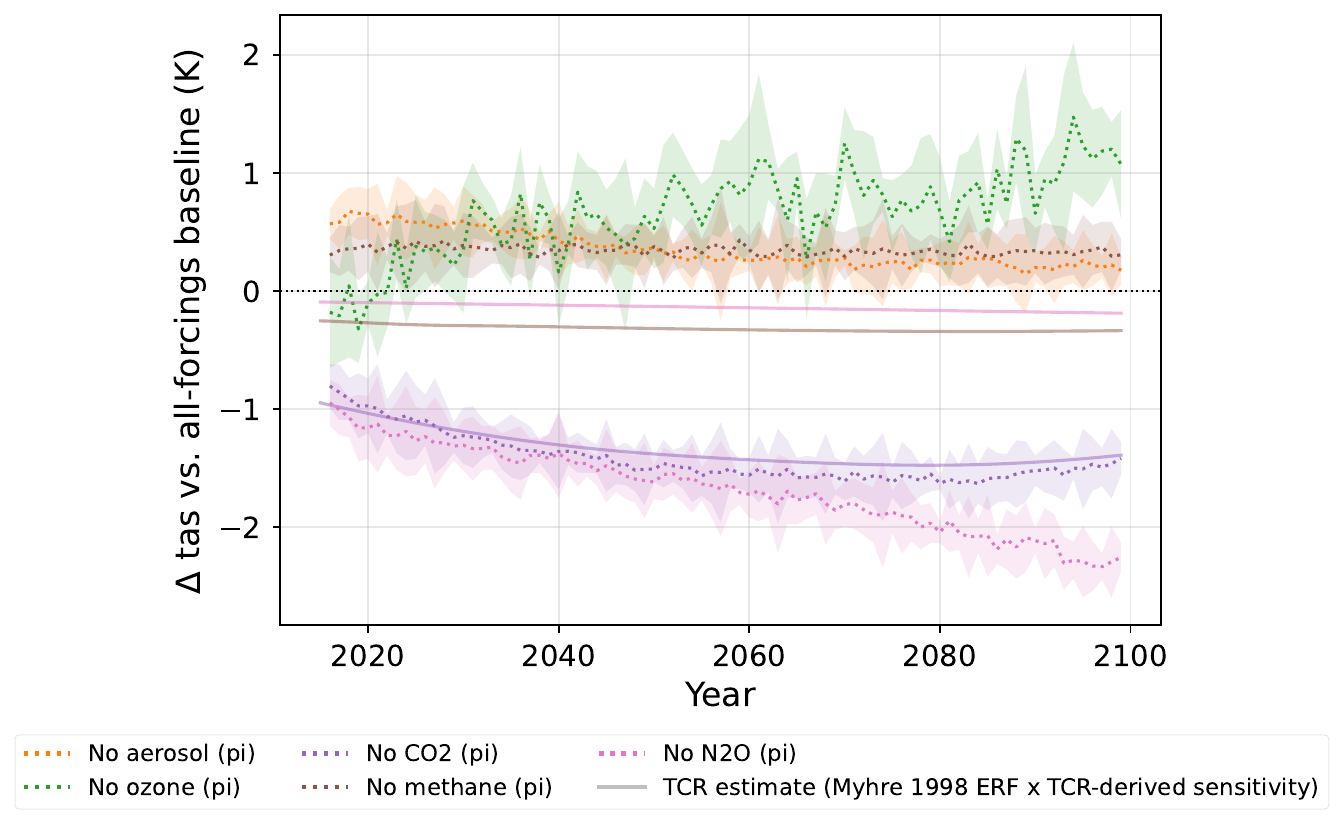}
\caption{$p_{\text{all}}=1.0$ (Single-forcing piControl-held ablations under SSP4-3.4, same methodology as \autoref{fig:surface_forced_response}. At $p_{\text{all}}=1.0$ (never trained with forcing-dropout), methane's ablation trace has the \emph{wrong sign} relative to its TCR reference, and N$_2$O's over-attribution is worse than at $p_{\text{all}}=0.8$ -- always seeing the full forcing set during training leaves the model's single-forcing-ablation behavior less reliable at inference, even though aggregate SSP4-3.4 accuracy is similar between the two (\autoref{sec:forcing_dropout_search}).}
\label{fig:w050_100_20_434_forced_response}
\end{figure}

\autoref{sec:forcing_dropout_search} finds $p_{\text{all}}=0.8$ and $p_{\text{all}}=1.0$ close to each other on aggregate SSP4-3.4 accuracy, both markedly better than lower $p_{\text{all}}$ values. Here we compare their single-forcing piControl-held attribution behaviour directly, SSP4-3.4, using the TCR-derived-sensitivity reference as \autoref{fig:surface_forced_response}. Despite their similar aggregate accuracy, the two differ noticeably in attribution behaviour (\autoref{fig:w050_80_10_434_forced_response}, \autoref{fig:w050_100_20_434_forced_response}): at $p_{\text{all}}=1.0$, methane's ablation response has the wrong sign relative to its TCR reference (positive rather than the expected negative), and N$_2$O's over-attribution is more severe than at $p_{\text{all}}=0.8$. Since $p_{\text{all}}=1.0$ means the model never trains with any forcing masked, we read this as evidence that some exposure to forcing-dropout during training, not just high aggregate accuracy on the fully-forced case, is needed for reliable single-forcing attribution behaviour at inference.

\subsection{Ensemble Size, Spread, and Skill (Energy-Score Model)}\label{app:ensemble_size_energy_score}

We report ensemble spread and RMSE for several ensemble sizes evaluated globally on SSP4-3.4 (\texttt{val}) with a much larger, 20-independently-seeded-rollout ensemble. We report both RMSE of the ensemble-mean prediction and the spread-skill ratio (SSR), defined as $\mathrm{SSR} = \sqrt{\frac{M+1}{M}\,\overline{\mathrm{Var}_M}} \,/\, \mathrm{RMSE}$, where $\overline{\mathrm{Var}_M}$ is the finite-sample-corrected ensemble variance averaged over all grid cells and the full rollout, and $M$ is the ensemble size; a perfectly calibrated ensemble has $\mathrm{SSR}=1$, values below 1 indicate an under-dispersed (overconfident) ensemble, and values above 1 indicate an over-dispersed ensemble.

\begin{table}[ht]
\centering
\begin{tabular}{lccc}
\toprule
\textbf{Ensemble size $M$} & \textbf{SSR} & \textbf{RMSE (K)} & \textbf{Spread (K)} \\
\midrule
1  & n/a   & 2.525 & n/a \\
5  & 1.012 & 2.544 & 2.576 \\
10 & 1.048 & 2.432 & 2.550 \\
15 & 1.071 & 2.405 & 2.576 \\
20 & 1.077 & 2.388 & 2.573 \\
\bottomrule
\end{tabular}
\caption{Global (area-weighted) $tas$ SSR and RMSE of the ensemble-mean prediction as a function of ensemble size, for AC-SSP, SSP4-3.4 (val), full 80-year rollout, 20 independently-seeded rollouts of the same checkpoint. RMSE at $M=1$ is not a meaningful spread comparison (a single member has no ensemble variance), so SSR/spread are reported as n/a there.}
\label{tab:pf8_ensemble_size}
\end{table}

This energy-score deterministic model's ensemble is close to well-calibrated globally, and calibration steadily decreases as members are added rather than plateauing early: SSR rises from 1.012 at $M=2$ to 1.077 at $M=20$, crossing the ideal value of 1.0 somewhere around $M=5$ and drifting to a mild over-dispersion (SSR $\approx 1.05$-$1.08$) beyond about 10 members. RMSE improves monotonically and substantially with ensemble size, from 2.680K at $M=2$ down to 2.388K at $M=20$, continuing to fall noticeably even past 10 members rather than saturating. This steady RMSE improvement indicates that a substantial share of this model's error is genuine sampling noise from the energy-score stochastic rollout rather than a shared bias, so larger ensembles are an effective way to reduce error for this model.

\subsection{Energy-Score Noise Scale: Effect on Calibration and Skill}\label{app:noise_scale}

The energy-score training objective (\cref{sec:energy_score_loss}) draws its stochasticity from Gaussian noise injected at inference time. Because this knob directly rescales the spread of the per-member stochastic perturbation without retraining, it offers a cheap way to trade ensemble spread against ensemble-mean RMSE. \autoref{tab:noise_scale_ssr} reports SSR and RMSE of the 5-member ensemble-mean prediction for AC-SSP as the noise level is swept over $\{1.0, 0.8, 0.6, 0.5, 0.4\}$, on SSP4-3.4 (val), full 85-year rollout.

\begin{table}[ht]
\centering
\begin{tabular}{cccc}
\toprule
\textbf{Noise scale} & \textbf{SSR} & \textbf{RMSE (K)} & \textbf{Spread (K)} \\
\midrule
1.0 & 1.012 & 2.544 & 2.576 \\
0.8 & 0.913 & 1.965 & 1.794 \\
0.6 & 0.711 & 1.967 & 1.399 \\
0.5 & 0.601 & 1.972 & 1.185 \\
0.4 & 0.485 & 1.981 & 0.961 \\
\bottomrule
\end{tabular}
\caption{Global (area-weighted) $tas$ SSR and RMSE of the 5-member ensemble-mean prediction as a function of \texttt{inference.energy\_score\_noise\_scale}, for AC-SSP, full 85-year rollout.}
\label{tab:noise_scale_ssr}
\end{table}

Reducing the noise scale below 1.0 pushes SSR steadily downward as well as RMSE across the whole sweep. In this configuration, then, noise scale is a trade-off between variance and accuracy: shrinking it trades away calibration for RMSE gains.

\subsection{Flow Matching Performance of the Forced Response}\label{app:flow_destroys_signal}

\begin{figure}\includegraphics[width=1\textwidth]{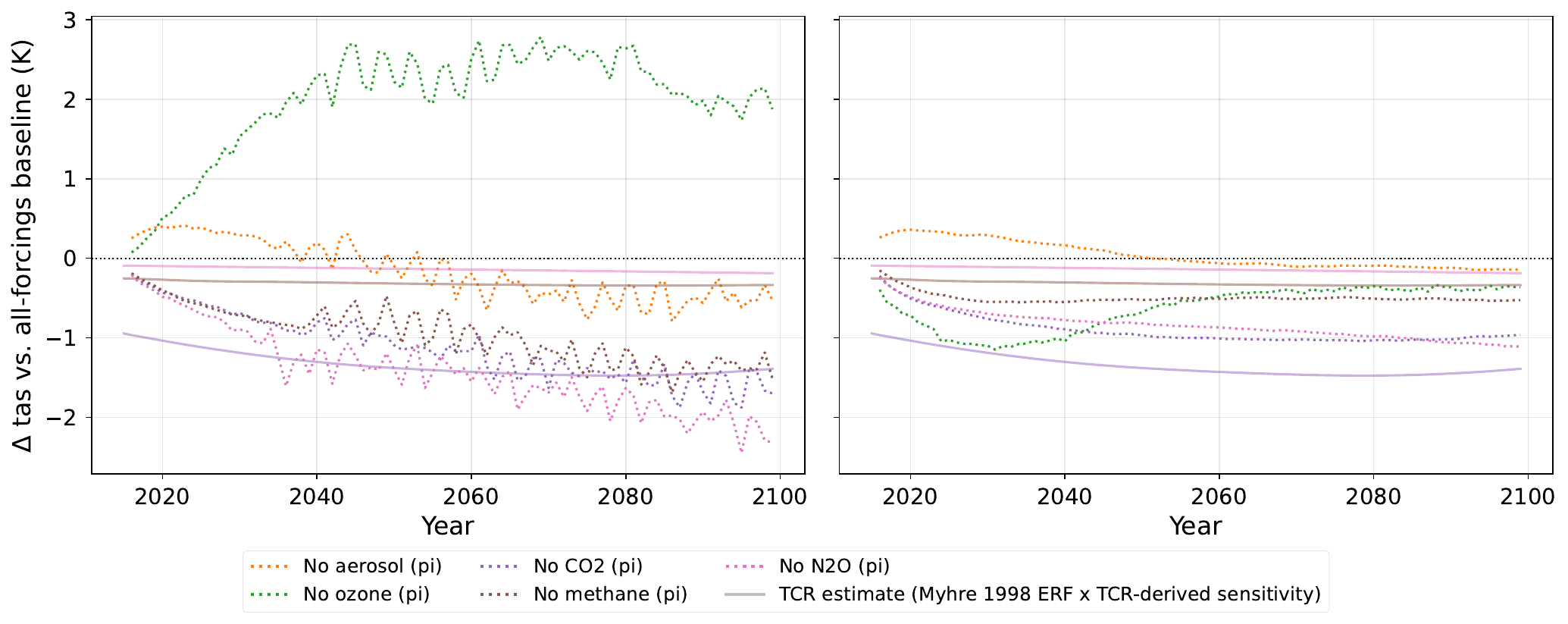}\caption{Single-forcing piControl-held ablations for SSP4-3.4: annual global-mean $tas$ deviation from each model's own all-forcings baseline, for all five single-forcing ablations. The flow-matching model (left) is compared against its underlying deterministic base model (right). For CO$_2$/methane/N$_2$O, the same-colored solid, semi-transparent line is that gas's TCR-derived sensitivity reference (same convention as \autoref{fig:surface_forced_response}); no aerosol/ozone reference is shown, as the simple ERF formulas used here do not cover those forcings. The deterministic base model's ablation traces evolve smoothly and track their TCR references, while the flow-matching model's traces are dominated by high-frequency, step-to-step noise that obscures the underlying forced-response signal, motivating the switch to the energy-score objective.} \label{fig:flow_damip_pf8_fixed_v2_forced_response}\end{figure}

In preliminary investigations, we adopted the residual flow-matching paradigm proposed by \cite{clyneArchesClimateProbabilisticDecadal2025,couaironArchesWeatherGenSkillfulComputeefficient2026}. This involved training a deterministic base model (utilizing the identical forcing dropout and pushforward mechanisms described in our current energy-score configuration) and subsequently training a flow-matching model to predict its step-to-step residuals. However, this approach yielded a marked degradation in the fidelity of the forced response, as illustrated in \autoref{fig:flow_damip_pf8_fixed_v2_forced_response}.

Applying our piControl-held ablation methodology to both architectures reveals a stark contrast. The deterministic base model (right column) produces traces that align closely with their transient climate response (TCR) references, evolving smoothly throughout the rollout. Conversely, the flow-matching model (left column) introduces pronounced high-frequency, step-to-step noise. While it tracks the broad underlying trend without runaway divergence—indicating that the forced-response signal is partially preserved rather than entirely obliterated—the variance is severely inflated. Notably, the no-aerosol and no-ozone ablations exhibit the largest absolute deviations in both architectures; the flow model's no-ozone trace, in particular, diverges by approximately 2.0 to 2.8~K by mid-century before stabilizing.We attribute this degradation to a misalignment of optimization objectives. The single-step flow-matching objective aggressively fits high-frequency, stochastic local variability, effectively overriding the low-amplitude, long-term forcing signal that the pushforward mechanism successfully embedded in the deterministic base model. Because extending flow matching to multi-step (pushforward) fine-tuning is computationally prohibitive, correcting this imbalance within the flow paradigm was unfeasible. We therefore required a generative framework natively compatible with multi-step optimization. Although recent literature has demonstrated multi-step fine-tuning via consistency models \cite{stockSwiftAutoregressiveConsistency2025b}, our attempts to distill this specific flow model into a consistency model proved unsuccessful, ultimately necessitating the architectural pivot to the energy-score approach detailed in the main text.

\subsection{Metrics}\label{app:metrics}

This section defines the diagnostic metrics used to evaluate model performance, variability, and uncertainty. Throughout these definitions, $Y$ denotes the target (ground truth) values and $\hat{Y}$ denotes the predicted values.

\textbf{Latitude-Weighted RMSE} To account for the decreasing surface area of grid cells toward the poles, we define the Latitude-Weighted Root Mean Square Error (RMSE) for an ensemble member $e$ as:\begin{equation}\text{RMSE}_e = \sqrt{ \frac{\sum_{t, \phi, \lambda} \cos(\phi) \cdot (Y_{e,t,\phi,\lambda} - \hat{Y}_{e,t,\phi,\lambda})^2}{\sum_{t, \phi, \lambda} \cos(\phi)} }\end{equation}where $\phi$ is the latitude, $\lambda$ is the longitude, and $t$ is the timestep. This value is then averaged across ensemble members. 

\textbf{NRMSE} To allow for meaningful comparisons across variables with varying scales, we utilize the Normalized RMSE (NRMSE). This is calculated by applying the Latitude-Weighted RMSE to variables that have been pre-normalized by their respective standard deviations. 

\textbf{Interannual Variability} (IAV) The grid-point IAV measures the magnitude of year-to-year fluctuations after removing the long-term forced trend. For an ensemble member $e$, it is defined as:\begin{equation}IAV_e(\phi, \lambda) = \sqrt{ \frac{1}{N_t - 1} \sum_{t=1}^{N_t} \left( \hat{Y}_{e,t,\phi,\lambda} - \mathcal{ls}_{e,\phi,\lambda}(t) \right)^2 }\end{equation}where $\mathcal{ls}_{e,\phi,\lambda}(t)$ represents the least-squares linear fit of the time series at that specific grid point and time. Each IAV calculation is independently detrended to prevent long-term warming trends from conflating the assessment of annual variability. IAV is based on temperature anomalies relative to a 1994–2014 climatological baseline.

\textbf{Ensemble Spread} The ensemble spread quantifies the uncertainty between different model realizations. First, we calculate the global spatial average for each ensemble member $e$ at timestep $t$:\begin{equation}\bar{Y}{e,t} = \frac{1}{N{\phi} N_{\lambda}} \sum_{\phi=1}^{N_{\phi}} \sum_{\lambda=1}^{N_{\lambda}} \hat{Y}_{e,t,\phi,\lambda}\end{equation}
Next, we compute the standard deviation across ensemble members ($N_E$) to find the instantaneous spread $\sigma_t$:\begin{equation}\sigma_t = \sqrt{\frac{1}{N_E - 1} \sum_{e=1}^{N_E} \left( \bar{Y}_{e,t} - \langle \bar{Y}_{\cdot,t} \rangle \right)^2}\end{equation}where $\langle \bar{Y}_{\cdot,t} \rangle$ is the ensemble mean of the spatial averages. The final Average Ensemble Spread is the temporal mean of these values: $\frac{1}{N_t} \sum_{t} \sigma_t$. 

\textbf{Spatial Standard Deviation} To evaluate the model's ability to capture spatial gradients, we calculate the standard deviation over the spatial dimensions ($\phi, \lambda$) for each timestep, subsequently averaged over time:\begin{equation}\sigma_{spatial} = \frac{1}{N_t} \sum_{t=1}^{N_t} \left( \sqrt{\frac{1}{N_\phi N_\lambda - 1} \sum_{\phi, \lambda} (\hat{Y}_{t,\phi,\lambda} - \bar{Y}_{t})^2} \right)\end{equation}

\textbf{Hydrostatic Balance}
We compare two different ways of calculating the thickness of atmospheric layers in order to understand if the model is learning a good relationship between $ta$ and $zg$. First, we check the Geometric Thickness, which is the direct difference between the geopotential height ($zg$) values at two adjacent pressure levels. Next, we check the Hydrostatic Thickness. This is the thickness calculated using the temperature ($ta$) and the known pressure levels. If the generated states are physically consistent, these two values should be nearly identical.

The fundamental relationship being checked is derived from the hydrostatic equation and the ideal gas law: \begin{equation}\Delta z = \frac{R_d \bar{T}}{g} \ln \left( \frac{p_{lower}}{p_{upper}} \right)\end{equation} 
Where: $R_d$ is the gas constant for dry air ($287.05 \text{ J kg}^{-1} \text{ K}^{-1}$). $g$ is the gravitational acceleration ($9.80665 \text{ m s}^{-2}$). $\bar{T}$ is the mean temperature of the layer. $p$ is the pressure.

Layer Thickness from Geopotential ($\Delta z_{geo}$): This is calculated as the difference between adjacent levels in the geopotential height array $z_g$: \begin{equation}\Delta z_{geo} = z_{g, i+1} - z_{g, i}\end{equation}

Average Temperature ($\bar{T}$): The mean layer temperature using the arithmetic mean of the temperature $T$ at the boundaries:
\begin{equation}\bar{T} = \frac{T_i + T_{i+1}}{2}\end{equation}

Log-Pressure Difference ($\Delta \ln p$): The difference in natural logarithms of pressure:
\begin{equation}\Delta \ln p = \ln\left(\frac{p_i}{p_{i+1}}\right)\end{equation}

Hydrostatic Thickness ($\Delta z_{hydro}$): Combining these, the theoretical thickness based on the hydrostatic assumption is:
\begin{equation}\Delta z_{hydro} = \left( \frac{R_d \bar{T}}{g} \right) \Delta \ln p\end{equation}

Finally, we calculate the residual (the error) and the Mean Absolute Error (MAE):
\begin{equation} \varepsilon = \Delta z_{geo} - \Delta z_{hydro}\end{equation}

Mean Absolute Error (MAE):\begin{equation}\text{MAE} = \frac{1}{N} \sum_{n=1}^{N} |\varepsilon_n|\end{equation}

\subsection{Energy Balance}\label{app:energy_balance}

Global energy conservation is one of the few constraints that must hold for
any closed representation of the climate system regardless of resolution or
parameterization choice, and it is the standard lens through which the
observed climate system's radiative and heat-storage terms are reconciled
\citep{trenberthImperativeClimateChange2009,vonschuckmannHeatStoredEarth2020}. We adapt the same bookkeeping to
check whether the rate of change of the emulator's modeled energy reservoirs
(atmosphere and upper ocean) is consistent with its own net surface flux.

\subsubsection{Column atmospheric energy}

The column-integrated atmospheric energy per unit area combines internal
(sensible), potential, kinetic, and latent contributions,

\begin{equation}
E_{\mathrm{atm}} = \int_0^{ps}
  \Big( c_p\, ta + g\, zg + \tfrac{1}{2}(ua^2+va^2) + L_v\, hus \Big)
  \frac{dp}{g},
\label{eq:atm-energy}
\end{equation}

with $c_p = 1004.6\ \mathrm{J\,kg^{-1}K^{-1}}$ and
$L_v = 2.5\times10^{6}\ \mathrm{J\,kg^{-1}}$. This is from equation 3.14 in (https://sci-hub.fr/10.1103/RevModPhys.56.365). We discretize
Eq.~\ref{eq:atm-energy} on the fixed pressure grid by assigning each level $i$
a layer $[\,u_i,\, \ell_i\,]$ bounded by the midpoints to its neighbors
($u_i = (p_{i-1}+p_i)/2$, $\ell_i = (p_i+p_{i+1})/2$, with $u_1 = 0$ at the
model top and $\ell_n \to \infty$ at the surface-most level), and clip each
layer to the actual surface pressure of the grid cell:

\begin{equation}
\Delta p_i = \max\big(0,\ \min(\ell_i, ps) - u_i\big),
\qquad
E_{\mathrm{atm}} \approx \sum_i \frac{e_i\, \Delta p_i}{g},
\label{eq:atm-energy-discrete}
\end{equation}

where $e_i$ is the bracketed integrand of Eq.~\ref{eq:atm-energy} evaluated at
level $i$. This clipping is necessary because $ps$ varies with orography and
surface pressure while the pressure levels are fixed: levels that lie below
the actual surface ($u_i \ge ps$) are excluded, and the lowest valid layer is
truncated at $ps$ rather than at its nominal lower bound, following the
standard treatment of terrain-following mass integrals in fixed-pressure-level
archives.

\subsubsection{Ocean heat content}

The upper-ocean heat content per unit area is the depth integral

\begin{equation}
E_{\mathrm{oce}} = \int_0^{-H} \rho_w c_{pw}\, thetao\; dz,
\label{eq:ocean-energy}
\end{equation}

with $\rho_w = 1026\ \mathrm{kg\,m^{-3}}$ and
$c_{pw} = 3991\ \mathrm{J\,kg^{-1}K^{-1}}$ \citep{levitusWorldOceanHeat2012}, seawater density and seawater heat capacity \cite{griffiesOMIPContributionCMIP62016} discretized
analogously to Eq.~\ref{eq:atm-energy-discrete} using layer thicknesses
$\Delta z_i$ derived from the midpoints between successive depth levels
$z_1, \dots, z_m$ ($H = 147.4$ m here).

\subsubsection{Global integration and the conservation check}

Column energies and $\mathrm{net\_flux}$ are integrated over the sphere using
the area element

\begin{equation}
A(\phi) = R_\oplus^2 \cos\phi\, \Delta\phi\, \Delta\lambda,
\label{eq:area}
\end{equation}

with $R_\oplus = 6.371\times10^6$ m, on the model's regular
$144\times144$ latitude--longitude grid. Writing
$\mathbf{E}_{\mathrm{atm}}(t) = \sum_{i,j} E_{\mathrm{atm}}(t,i,j)\,A_{i,j}$
(and analogously for $\mathbf{E}_{\mathrm{oce}}$ and
$\mathbf{F}_{\mathrm{net}}$), conservation of energy at monthly resolution
requires

\begin{equation}
\frac{\Delta\mathbf{E}_{\mathrm{atm}} + \Delta\mathbf{E}_{\mathrm{oce}}}{\Delta t}
    \;\approx\; \overline{\mathbf{F}_{\mathrm{net}}},
\label{eq:conservation}
\end{equation}

where $\Delta t$ is taken as the average calendar month
($365.25/12$ days) and $\overline{\mathbf{F}_{\mathrm{net}}}$ is the mean of
$\mathbf{F}_{\mathrm{net}}$ over the two months bounding $\Delta t$. The
residual of Eq.~\ref{eq:conservation} is reported both as a global time series
and, for the trailing decade of a rollout, as a spatial map of
$\langle E_{\mathrm{atm}}\rangle$ and $\langle E_{\mathrm{oce}}\rangle$ for the
emulator, the reference, and their difference, which localizes systematic
bias rather than only detecting it in the global mean.

\subsection{Effective Radiative Forcing and Transient Climate Response Analysis}\label{app:tcr_explanation}

For the greenhouse gases (CO$_2$, CH$_4$, N$_2$O), we build a physically expected $\Delta tas$ reference for each ablation by combining two standard quantities, Effective Radiative Forcing (ERF) and Transient Climate Response (TCR), in three steps.

We convert each gas's concentration trajectory $C(t)$ into an effective radiative forcing $\mathrm{ERF}(t)$ in W/m$^2$, relative to its pre-industrial reference concentration $C_0$ (CO$_2$: $278$~ppm, CH$_4$: $700$~ppb, N$_2$O: $270$~ppb). For CO$_2$ we use the standard closed-form expression of \cite{myhreNewEstimatesRadiative1998}:
\begin{equation}
\mathrm{ERF}_{\mathrm{CO}_2}(C) = 5.35 \ln(C / C_0).
\end{equation}
For CH$_4$ and N$_2$O we instead use the expressions of \cite{etminanRadiativeForcingCarbon2016} which additionally capture the cross-sensitivity of each gas's ERF to the other's concentration and to CO$_2$:
\begin{align}
\mathrm{ERF}_{\mathrm{CH}_4}(M) &= \left(a_3 \bar{M} + b_3 \bar{N} + 0.043\right)\left(\sqrt{M} - \sqrt{M_0}\right) \\
\mathrm{ERF}_{\mathrm{N}_2\mathrm{O}}(N) &= \left(a_2 \bar{C} + b_2 \bar{N} + c_2 \bar{M} + 0.117\right)\left(\sqrt{N} - \sqrt{N_0}\right)
\end{align}
where $a_3 = -1.3\times10^{-6}$, $b_3 = -8.2\times10^{-6}$, $a_2 = -8.0\times10^{-6}$, $b_2 = 4.2\times10^{-6}$, and $c_2 = -4.9\times10^{-6}$ (all in W/m$^2$ per the relevant concentration unit, ppm for CO$_2$ and ppb for CH$_4$/N$_2$O); the other two gases are held at their own current-trajectory values (not pre-industrial) in each formula, following \cite{etminanRadiativeForcingCarbon2016}.

ERF alone is a unit of energy flux and turning it into an expected $\Delta tas$ requires a climate sensitivity. We use TCR (Transient Climate Response: the transient warming expected at CO$_2$ doubling under a 1\%/yr ramp, \cite{gregoryTransientClimateResponse2008}). Rather than a single generic TCR value, we use a model-specific estimate for each source ESM: $\mathrm{TCR} = 2.45$K for IPSL-CM6A-LR and $\mathrm{TCR} = 2.8$K for CanESM5 \cite{boucherPresentationEvaluationIPSLCM6ALR2020,swartCanadianEarthSystem2019a}. Because TCR is defined per unit of CO$_2$-doubling forcing, we convert it to a sensitivity in K per W/m$^2$ by dividing by the forcing at CO$_2$ doubling, $F_{2\times\mathrm{CO}_2} = 5.35\ln 2 \approx 3.71$W/m$^2$ (the same constant appearing in the CO$_2$ ERF formula above, evaluated at $C/C_0=2$):
\begin{equation}
\lambda_{\text{TCR}} = \mathrm{TCR} / F_{2\times\mathrm{CO}_2}.
\end{equation}

Multiplying a gas's ERF trajectory by $\lambda_{\text{TCR}}$ converts it into the physically expected transient temperature response to that gas alone, under the linear-response assumption TCR itself already embeds:
\begin{equation}
\Delta tas_{\text{expected}}(t) = -\lambda_{\text{TCR}} \cdot \mathrm{ERF}_{\text{gas}}\!\left(C_{\text{gas}}(t)\right),
\end{equation}
where the sign is negative because setting a gas to its pre-industrial control level removes its forcing relative to the all-forcings baseline, so a gas with positive (warming) ERF is expected to produce cooling once held at pi-control. This is the solid reference trace plotted against each model's own piControl-held ablation trajectory (dotted) seen throughout this paper.

Unlike a per-gas TCR fit to a specific single-forcing experiment (only available from the INCA-driven IPSL DAMIP suite), the CH$_4$/N$_2$O ERF$\times$TCR reference above needs only each gas's own concentration trajectory (available for any model from the same input4MIPs forcing files used at training time) and the host model's own overall TCR; we therefore construct it identically for both IPSL-CM6A-LR and CanESM5, using \cite{etminanRadiativeForcingCarbon2016}'s CH$_4$/N$_2$O ERF formulas and each model's own TCR ($2.45$K for IPSL-CM6A-LR, $2.8$K for CanESM5 \cite{,boucherPresentationEvaluationIPSLCM6ALR2020,swartCanadianEarthSystem2019a}).

\subsection{CanESM5 vs.\ IPSL Training Data}\label{app:canesm5_ipsl_members}

Here we explain the data and procressing for the CanESM5 data, as well as a comparison to available data of the IPSL. We retain CanESM5 in its native 64×128 grid, in contrast to the 144×144 configuration of the IPSL model. Consequently, we apply an altered window size in the SWIN component ([8, 16] rather than [6, 10]). CanESM5 provides 19 pressure levels, which constitutes a superset of the 17 levels present in IPSL. Both models share the [100000, ..., 1000] Pa range, but CanESM5 includes two additional uppermost levels (500~Pa and 100~Pa) that we cut. For the ocean temperature field (\textit{thetao}), CanESM5 uses 45 levels beginning at the surface, whereas IPSL employs 10 levels starting at a depth of 0.51 m. To maintain layer-integrated heat content across these differing discretizations, we project the CanESM5 data onto the IPSL target depths using a conservative, bounds-overlap-weighted vertical remapping. The derivation of forcing fields remains identical across both models, except for the aerosol forcings. Because the aerosol forcing in CanESM5 is structured differently from the INCA-derived fields used for IPSL (detailed in \autoref{sec:forcings}), we omit the aerosol forcing for CanESM5.

The datasets also exhibit discrepancies in ensemble size, as the number of realizations for CanESM5 is smaller than that of IPSL. \autoref{tab:canesm5_ipsl_members} (\autoref{app:canesm5_ipsl_members}) shows the realizations utilized for each shared training experiment. Additionally, two specific scenarios present in the IPSL training, SSP1-1.9 and SSP4-6.0, lack CanESM5 equivalents within our dataset.
\autoref{tab:canesm5_ipsl_members} gives the per-experiment realization counts referenced in \autoref{sec:canesm5_application}.

\begin{table}[ht]
\centering
\begin{tabular}{lcc}
\toprule
\textbf{Experiment} & \textbf{IPSL realizations} & \textbf{CanESM5 realizations} \\
\midrule
historical & 10 & 10 \\
piControl & 2 & 0 (excluded) \\
abrupt-4xCO2 & 1 & 0 (excluded) \\
SSP1-1.9 & 6 & 0 (excluded) \\
SSP1-2.6 & 6 & 3 \\
SSP2-4.5 & 6 & 5 \\
SSP3-7.0 & 10 & 9 \\
SSP4-3.4 (val target) & 1 & 1 \\
SSP4-6.0 & 6 & 0 (excluded) \\
SSP5-3.4-over (val3 target) & 1 & 1 \\
SSP5-8.5 (test target) & 6 & 2 \\
\bottomrule
\end{tabular}
\caption{Number of ensemble-member realizations (CMIP6 \textit{r*i1p1f1} variants) used per training/evaluation experiment, IPSL vs.\ CanESM5. Counts reflect what our processed archive actually contains, not necessarily the full set nominally available from each model's CMIP6 submission. CanESM5's historical realizations are a non-contiguous subset (r1, r2, r3, r5, r7, r10, r11, r12, r13, r14, r16, r18, r19, r20, r22, r24, r25), unlike IPSL's contiguous r1-r10.}\label{tab:canesm5_ipsl_members}
\end{table}

\subsection{Rest of Model Performance} \label{app:rest_of_model_performance}

For completeness, \autoref{tab:atm_level_performance} and \autoref{tab:ocean_depth_performance} report per-level performance for every remaining atmospheric and ocean variable not otherwise broken out elsewhere in this paper: last-decade global RMSE against the CMIP target, spatial standard deviation (SpStd), and detrended inter-annual variability (IAV), for $AC\text{-}SSP$ on SSP5-3.4 with a 5-member ensemble. SpStd and IAV cells are given as \textit{model (IPSL)}, so the model's own value can be compared directly against the IPSL target's at the same level. Spatial std and IAV are computed identically to the whole-column ($tas$) diagnostics used elsewhere in this paper (\autoref{app:ensemble_size_energy_score}), applied per pressure/depth level over the same final-decade window as the RMSE figures.

\begin{table}[ht]
\centering
\resizebox{\textwidth}{!}{%
\begin{tabular}{lcccccc}
\toprule
 & \multicolumn{3}{c}{ta [K]} & \multicolumn{3}{c}{zg [m]} \\
Pressure level (hPa) & RMSE & SpStd & IAV & RMSE & SpStd & IAV \\
\midrule
1000 & 0.373 & 11.6 (12.4) & 0.441 (0.463) & 4.04 & 53.1 (56.2) & 5.28 (6.78) \\
925 & 0.307 & 11.4 (11.5) & 0.513 (0.442) & 3.17 & 95.7 (96.1) & 6.98 (8.54) \\
850 & 0.327 & 10.9 (10.9) & 0.499 (0.436) & 3.09 & 117 (117) & 7.52 (8.62) \\
700 & 0.25 & 12.5 (12.6) & 0.443 (0.429) & 3.61 & 177 (176) & 8.35 (10.7) \\
600 & 0.25 & 11.5 (11.5) & 0.444 (0.403) & 4.19 & 233 (232) & 8.62 (12) \\
500 & 0.216 & 11.8 (11.7) & 0.458 (0.399) & 4.55 & 293 (292) & 10.8 (13.7) \\
400 & 0.219 & 11.8 (11.8) & 0.378 (0.394) & 6.77 & 368 (366) & 11.8 (15.7) \\
300 & 0.221 & 10.8 (10.7) & 0.297 (0.369) & 8.51 & 463 (461) & 14.5 (18.2) \\
250 & 0.183 & 9.03 (8.98) & 0.34 (0.399) & 8.45 & 516 (513) & 16 (19.4) \\
200 & 0.233 & 5.31 (5.33) & 0.532 (0.543) & 8.86 & 561 (558) & 15.6 (20.4) \\
150 & 0.271 & 4.09 (4.19) & 0.564 (0.636) & 9.55 & 573 (571) & 15.4 (21.7) \\
100 & 0.271 & 9.33 (9.3) & 0.496 (0.582) & 11.5 & 528 (526) & 15.1 (24.5) \\
70 & 0.302 & 8.38 (8.36) & 0.502 (0.661) & 11.8 & 478 (479) & 15.9 (26.6) \\
50 & 0.32 & 5.57 (5.62) & 0.491 (0.678) & 11.3 & 462 (465) & 18.6 (28.7) \\
30 & 0.312 & 3.52 (3.74) & 0.453 (0.617) & 13.2 & 480 (487) & 22.2 (33.1) \\
20 & 0.3 & 3.07 (3.35) & 0.456 (0.627) & 16.2 & 511 (520) & 25.4 (36.4) \\
10 & 0.3 & 2.41 (2.62) & 0.483 (0.586) & 18.3 & 559 (574) & 28.1 (41.2) \\
\bottomrule
\end{tabular}%
}
\caption{$AC\text{-}SSP$ atmospheric-column performance by pressure level, SSP5-3.4-over, last decade of the rollout, for $ta$ and $zg$. SpStd/IAV given as model (IPSL). Continued for $hus$/$ua$ in \autoref{tab:atm_level_performance_2} and $va$ in \autoref{tab:atm_level_performance_3}.} \label{tab:atm_level_performance}
\end{table}

\begin{table}[ht]
\centering
\resizebox{\textwidth}{!}{%
\begin{tabular}{lcccccc}
\toprule
 & \multicolumn{3}{c}{hus [kg/kg]} & \multicolumn{3}{c}{ua [m/s]} \\
Pressure level (hPa) & RMSE & SpStd & IAV & RMSE & SpStd & IAV \\
\midrule
1000 & 0.00013 & 0.00538 (0.00544) & 0.000205 (0.00022) & 0.199 & 3.97 (3.94) & 0.445 (0.535) \\
925 & 0.000121 & 0.00477 (0.00475) & 0.000187 (0.00019) & 0.246 & 5.11 (5.12) & 0.64 (0.666) \\
850 & 0.000103 & 0.00386 (0.00385) & 0.000175 (0.000173) & 0.269 & 5.24 (5.29) & 0.652 (0.713) \\
700 & 8.26e-05 & 0.00222 (0.00221) & 0.000128 (0.00014) & 0.337 & 5.84 (5.87) & 0.797 (0.843) \\
600 & 5.92e-05 & 0.00155 (0.00153) & 9.18e-05 (0.000109) & 0.349 & 6.39 (6.44) & 0.883 (0.937) \\
500 & 3.71e-05 & 0.000797 (0.000792) & 5.54e-05 (6.48e-05) & 0.414 & 7.23 (7.27) & 1.13 (1.09) \\
400 & 1.85e-05 & 0.000384 (0.00038) & 2.42e-05 (2.94e-05) & 0.483 & 8.47 (8.49) & 1.18 (1.3) \\
300 & 6.74e-06 & 0.000138 (0.000136) & 7.39e-06 (9.3e-06) & 0.542 & 10.4 (10.4) & 1.35 (1.51) \\
250 & 3.18e-06 & 6.54e-05 (6.42e-05) & 3.21e-06 (4.09e-06) & 0.567 & 11.7 (11.7) & 1.36 (1.59) \\
200 & 1.14e-06 & 2.33e-05 (2.28e-05) & 1.09e-06 (1.42e-06) & 0.559 & 12.9 (12.9) & 1.24 (1.58) \\
150 & 2.73e-07 & 5.34e-06 (5.21e-06) & 2.59e-07 (3.65e-07) & 0.505 & 12.8 (12.8) & 1.07 (1.49) \\
100 & 2.38e-08 & 4.04e-07 (3.93e-07) & 2.94e-08 (4.09e-08) & 0.465 & 10.3 (10.4) & 1.03 (1.4) \\
70 & 9.54e-09 & 7.89e-08 (7.32e-08) & 9.39e-09 (1.3e-08) & 0.532 & 10.3 (10.3) & 0.968 (1.32) \\
50 & 8.27e-09 & 7.43e-08 (7.17e-08) & 7.76e-09 (1.29e-08) & 0.66 & 11 (11.1) & 1.11 (1.69) \\
30 & 9.81e-09 & 4.03e-08 (3.58e-08) & 6.98e-09 (1.08e-08) & 0.813 & 12.3 (12.8) & 1.43 (2.11) \\
20 & 9.41e-09 & 1.42e-08 (7.47e-09) & 6.52e-09 (8.48e-09) & 0.863 & 13.7 (14.2) & 1.76 (2.41) \\
10 & 8.31e-09 & 8.72e-09 (4e-09) & 6.33e-09 (8.66e-09) & 1.96 & 14 (15.1) & 2.19 (3.3) \\
\bottomrule
\end{tabular}%
}
\caption{$AC\text{-}SSP$ atmospheric-column performance by pressure level, SSP5-3.4-over, last decade of the rollout, for $hus$ and $ua$. SpStd/IAV given as model (IPSL). Continued from \autoref{tab:atm_level_performance}.} \label{tab:atm_level_performance_2}
\end{table}

\begin{table}[ht]
\centering
\begin{tabular}{lccc}
\toprule
 & \multicolumn{3}{c}{va [m/s]} \\
Pressure level (hPa) & RMSE & SpStd & IAV \\
\midrule
1000 & 0.158 & 1.81 (1.82) & 0.375 (0.409) \\
925 & 0.187 & 1.9 (1.89) & 0.466 (0.459) \\
850 & 0.196 & 1.62 (1.63) & 0.46 (0.447) \\
700 & 0.212 & 1.77 (1.8) & 0.518 (0.52) \\
600 & 0.233 & 1.72 (1.75) & 0.58 (0.586) \\
500 & 0.275 & 1.67 (1.76) & 0.558 (0.689) \\
400 & 0.32 & 1.9 (1.93) & 0.781 (0.829) \\
300 & 0.373 & 2.22 (2.21) & 1.05 (0.977) \\
250 & 0.391 & 2.31 (2.38) & 0.957 (1.03) \\
200 & 0.397 & 2.34 (2.48) & 0.751 (1) \\
150 & 0.365 & 2.22 (2.37) & 0.6 (0.873) \\
100 & 0.29 & 1.9 (2.08) & 0.521 (0.709) \\
70 & 0.257 & 1.87 (2.1) & 0.451 (0.613) \\
50 & 0.264 & 2.12 (2.43) & 0.451 (0.618) \\
30 & 0.319 & 2.69 (3.08) & 0.504 (0.709) \\
20 & 0.367 & 3.14 (3.59) & 0.545 (0.817) \\
10 & 0.45 & 3.68 (4.18) & 0.624 (0.987) \\
\bottomrule
\end{tabular}
\caption{$AC\text{-}SSP$ atmospheric-column performance by pressure level, SSP5-3.4-over, last decade of the rollout, for $va$. SpStd/IAV given as model (IPSL). Continued from \autoref{tab:atm_level_performance_2}.} \label{tab:atm_level_performance_3}
\end{table}

\begin{table}[ht]
\centering
\small
\begin{tabular}{lccc}
\toprule
 & \multicolumn{3}{c}{thetao [K]} \\
Depth level (m) & RMSE & SpStd & IAV \\
\midrule
0.5 & 0.255 & 11.1 (11.2) & 0.259 (0.363) \\
3.9 & 0.253 & 11.1 (11.2) & 0.25 (0.36) \\
8.1 & 0.254 & 11.1 (11.2) & 0.243 (0.359) \\
14.0 & 0.264 & 11.2 (11.3) & 0.241 (0.358) \\
22.8 & 0.283 & 11.3 (11.3) & 0.245 (0.361) \\
35.7 & 0.309 & 11 (11) & 0.24 (0.363) \\
53.9 & 0.328 & 10.5 (10.5) & 0.238 (0.368) \\
77.6 & 0.348 & 9.8 (9.81) & 0.227 (0.343) \\
108.0 & 0.381 & 8.91 (8.91) & 0.211 (0.301) \\
147.4 & 0.402 & 7.91 (7.93) & 0.202 (0.27) \\
\bottomrule
\end{tabular}
\caption{$AC\text{-}SSP$ ocean-temperature ($thetao$) performance by depth level, SSP5-3.4-over, last decade of the rollout. SpStd/IAV given as model (IPSL).} \label{tab:ocean_depth_performance}
\end{table}

\ack{The authors are grateful to the CLEPS infrastructure from the Inria of Paris for providing resources and support. To process the data from IPSL, this study benefited from the IPSL
mesocenter ESPRI facility, which is supported by CNRS, UPMC, Labex L-IPSL, CNES and Ecole Polytechnique.}

\funding{G. Clyne and C. Monteleoni were primarily supported by the Choose France Chair in AI, from the French government.  }

\data{The code for this project is available at https://github.com/grahamclyne/ArchesClimate-SSP. The model and a data sample is available at https://huggingface.co/gclyne/ArchesClimate and https://huggingface.co/datasets/gclyne/ArchesClimate-data. With the exception of the aerosol forcings, the training data can be found at https://esgf-metagrid.cloud.dkrz.de/search.  }
\bibliographystyle{vancouver}

\bibliography{mlearth_submission}

\end{document}